\documentclass{article}
\usepackage{graphicx}

\title{Inferring the relationship between soil temperature and the normalized difference vegetation index with machine learning}

\author{Steven Mortier$^{a*}$, Amir Hamedpour$^{b,c}$, Bart Bussmann$^{a}$,\\ Ruth Phoebe Tchana Wandji$^{c}$, Steven Latr\'{e}$^{a}$, Bjarni D. Sigurdsson$^{c}$,\\ Tom De Schepper$^{a}$ and Tim Verdonck$^{d}$}

\date{}

\usepackage{amssymb}
\usepackage{mathtools}
\usepackage{hyphenat}
\usepackage{caption}
\usepackage{subcaption}
\usepackage{float}
\usepackage{booktabs}
\usepackage{enumitem}
\usepackage[hidelinks]{hyperref}
\usepackage[T1]{fontenc}
\usepackage{makecell} 
\usepackage{colortbl} 
\usepackage{acronym}
\usepackage[capitalize]{cleveref}
\usepackage{amsmath}
\usepackage{xcolor}
\usepackage[round]{natbib}

\acrodef{NDVI}[NDVI]{Normalized Difference Vegetation Index}
\acrodef{SOS}[SOS]{start of the season}
\acrodef{PEAK}[PEAK]{maximum annual NDVI value}
\acrodef{POS}[POS]{peak of the season}
\acrodef{ML}[ML]{machine learning}
\acrodef{ANN}[ANN]{artificial neural network}
\acrodef{MLP}[MLP]{multilayer perceptron}
\acrodef{MSE}[MSE]{mean squared error}
\acrodef{MAE}[MAE]{mean average error}
\acrodef{CV}[CV]{cross validation}
\acrodef{SHAP}[SHAP]{SHapley Additive exPlanations}
\acrodef{LIME}[LIME]{Local Interpretable Model-Agnostic Explanations}
\acrodef{xAI}[xAI]{explainable artificial intelligence}

\newcommand\blfootnote[1]{
  \begingroup
  \renewcommand\thefootnote{}\footnote{#1}
  \addtocounter{footnote}{-1}
  \endgroup
}

\DeclareMathOperator*{\argmax}{argmax}

\AtBeginDocument{\colorlet{defaultcolor}{.}} 
\definecolor{lightgray}{gray}{0.95}

\begin{document}

\maketitle

\begin{center}
$^a$University of Antwerp - imec, IDLab - Department of Computer
Science, Sint-Pietersvliet 7, 2000 Antwerp, Belgium\\
$^b$Svarmi, Data Company Specialized in Remote Sensing and Drones, Hlíðasmári 8, 201
Kópavogur, Iceland\\
$^c$Agricultural University of Iceland - AUI, Faculty of Environmental and Forest
Sciences, Hvanneyri, 311 Borgarnes, Iceland\\
$^d$University of Antwerp - imec, Department of Mathematics, \\Middelheimlaan 1, 2000
Antwerp, Belgium
\end{center}

\begin{center}
    \textbf{Abstract}
\end{center}
\blfootnote{$^*$ Corresponding author at steven.mortier@uantwerpen.be.}
Changes in climate can greatly affect the phenology of plants, which can have important feedback effects, such as altering the carbon cycle. These phenological feedback effects are often induced by a shift in the start or end dates of the growing season of plants. The normalized difference vegetation index (NDVI) serves as a straightforward indicator for assessing the presence of green vegetation and can also provide an estimation of the plants' growing season. In this study, we investigated the effect of soil temperature on the timing of the start of the season (SOS), timing of the peak of the season (POS), and the maximum annual NDVI value (PEAK) in subarctic grassland ecosystems between 2014 and 2019. We also explored the impact of other meteorological variables, including air temperature, precipitation, and irradiance, on the inter-annual variation in vegetation phenology. Using machine learning (ML) techniques and SHapley Additive exPlanations (SHAP) values, we analyzed the relative importance and contribution of each variable to the phenological predictions. Our results reveal a significant relationship between soil temperature and SOS and POS, indicating that higher soil temperatures lead to an earlier start and peak of the growing season. However, the Peak NDVI values showed just a slight increase with higher soil temperatures. The analysis of other meteorological variables demonstrated their impacts on the inter-annual variation of the vegetation phenology. Ultimately, this study contributes to our knowledge of the relationships between soil temperature, meteorological variables, and vegetation phenology, providing valuable insights for predicting vegetation phenology characteristics and managing subarctic grasslands in the face of climate change. Additionally, this work provides a solid foundation for future ML-based vegetation phenology studies.

\section{Introduction}
\label{sec:fa:introduction}

In-situ monitoring of changes in vegetation in inaccessible Arctic regions is challenging, prompting many such studies to rely on remote sensing techniques \citep{Zmarz2018ApplicationEcosystem}. In the field of remote sensing, vegetation indices such as the \ac{NDVI} are used to quantify and qualify vegetation cover \citep{Huang2021ASensing}. This is achieved through airborne or satellite spectral methods \citep{RYU2021SimpleSensor} or ground-level measurements, using handheld instruments \citep{Balzarolo2011Ground-BasedControversies,Ferrara2010ComparisonProperties}. Vegetation activity monitoring using \ac{NDVI} has shown both intra-annual and inter-annual  variations that can give valuable insights into ecosystem changes \citep{Beck2006ImprovedNDVI}. Some parameters that can be derived from such intra-annual seasonal \ac{NDVI} curves are the \ac{SOS}, \ac{POS}, and \ac{PEAK}. 

In northern latitudes, the intra-annual temperature and irradiance variation are important factors that control the cycles in the growth and reproduction of the flora \citep{Fenner1998ThePlants}. Over the last decades, different life-cycle events of vegetation (phenology) have been observed to change in this region  \citep{Epstein2013RecentVegetation}. This has been related to ongoing climate change \citep{IPCC2021TechnicalChange}, which has started to affect vegetation phenological cycles, productivity, and community structure \citep{Semenchuk2016HighPeriodicity}. Inter-annual analyses found relationships between climate change and these changes in vegetation dynamics, particularly with regard to the increase in surface temperature, resulting in an increased \ac{PEAK} \ac{NDVI} and with a notable impact on the length of the growing seasons \citep{Potter2020ChangesDecades, Arndt2019ArcticAlaska}. Starting from the year 2000, scientists started to name this phenomenon (the increase in \ac{PEAK}) ``Arctic greening''  \citep{Merrington2019AData}. This phenomenon was hypothesized to persist with continued climate warming, based on the compelling evidence of increased  \ac{PEAK} \ac{NDVI} \citep{Beck2011SatelliteDifferences}, plant productivity \citep{Loranty2012ShrubTundra}, phenology \citep{Semenchuk2016HighPeriodicity}, and vegetation composition \citep{Walker2012EnvironmentTransects} between 1980s and early 2000s \citep{Epstein2012DynamicsDecades,Epstein2013RecentVegetation}.

Interestingly, the ``Arctic greening'' effect has not occurred everywhere at high latitudes and since the early 2000s, the relationship between \ac{PEAK} \ac{NDVI} with an increase in surface temperature has weakened in many places \citep{Bhatt2013RecentTundra,Myers-Smith2020ComplexityArctic}. In fact, in some regions, this relationship has even become negative, introducing the term ``Arctic browning'' \citep{Beck2011SatelliteDifferences}. It is generally believed that the shift towards browning must indicate that other meteorological drivers (e.g., temperature, precipitation, wind, photoperiod) or biological drivers (e.g., insect grazing, drought, etc.) are in play. However, the issue still requires further study. 

In Iceland, the same strong ``Arctic greening'' trend was shown to occur during the 1980s-2000s as in many other high-latitude regions, but with a notable stagnation of the national \ac{PEAK} \ac{NDVI} during 2000-2010, even if the surface temperatures continued to increase in Iceland during that period \citep{Raynolds2015WarmingTrends,Bjornsson2007Loftslagsbreytingar2018}. What happened in Iceland after 2010 is unclear, but a recent study showed that the inter-annual variation in the national average \ac{PEAK} \ac{NDVI} has been large during 2001-2019 period \citep{Olafsson2021InfluenceSensing}. Therefore, it is of interest to further study how the \ac{NDVI} of Icelandic ecosystems responds to further warming. 

Continued climate change is expected to cause relatively higher increases in surface temperatures at higher latitudes in the coming decades \citep{IPCC2021TechnicalChange}, which will likely lead to relatively more ecosystem changes in plant productivity than at lower latitudes \citep{Chen2021BiophysicalArctic}. Potential changes include further temporal shifts in parameters that characterize growing seasons \citep{Semenchuk2016HighPeriodicity} and increases in plant productivity \citep{Street2022WhyLimited,VanDerWal2014High-arcticBiomass}. However, it is important to further investigate the warming impacts on \ac{NDVI} to better underpin such predictions for future changes. Combining data from manipulation (warming) experiments offer possibilities to study future high-latitude ecosystem \ac{NDVI} responses \citep{Bjorkman2020StatusMonitoring,Leblans2017PhenologicalWarming}.

To relate changes in vegetation composition, biomass or \ac{NDVI} to environmental parameters, traditional statistical methods like (non-)linear regression or linear mixed models have been most commonly used \citep{hope1993theRelationship,walker2012environment,Leblans2017PhenologicalWarming,estrella2021quantifying,wang2021Satellite}. Additionally, multivariate methods have also been used, for example multivariate analysis of variance tests \citep{Michielsen2014}. 

Despite massive advancements in the field of \ac{ML} during the last decade, \ac{ML} is not yet often used for vegetation studies.
\ac{ML} models can be used for various tasks, among which are classification, regression, and image segmentation. In \ac{ML}, models extract knowledge from data and use this knowledge to produce an output relevant to the task at hand. These models use three main learning paradigms: supervised learning, unsupervised learning or reinforcement learning. This study only considers the first paradigm, as we build a regression model. Within supervised learning, there are a multitude of model types, for example support vector machines \citep{Hearst1998SupportMachines}, boosted tree ensembles (e.g., XGBoost \citep{Chen2016XGBoost:System} or LightGBM \citep{Ke2017LightGBM:Tree}) and \acp{ANN} \citep{McCulloch1943AActivity}. This analysis will use \acp{ANN}, particularly \acp{MLP}, which are fully connected feedforward neural networks that consist of multiple layers of nodes that are connected with each other by weighted edges.

Recently, \ac{ML} has also shown promising results in the field of ecology \citep{Thessen2016AdoptionScience, Christin2019ApplicationsEcology}, for use cases such as species identification \citep{Barre2017LeafNet:Identification, Waldchen2018MachineIdentification, Chen2020AutomaticApproaches}, 
 behavioral studies \citep{Schofield2019ChimpanzeeLearning, Clapham2020AutomatedBears}, ecological modelling and forecasting \citep{Ye2011ForecastingNetwork, Cho2009CharacterizingMap, Strydom2021ATime}, remote sensing \citep{Li2020Deep-learningEcosystem, Guo2020ApplicationChallenges} and climate change studies \citep{Rolnick2022TacklingLearning, OGorman2018UsingEvents}, among others. The utilization of \ac{ML} techniques has opened new avenues for understanding complex ecological phenomena and predicting ecological responses. Considering the proven potential of \ac{ML} in addressing research questions in the field of ecology, we propose to apply \ac{ML} methods to investigate the relationship between vegetation phenology and environmental drivers in subarctic grasslands.

Unfortunately, \ac{MLP}s are black-box models. This means that, while they can approximate any function, it is nearly impossible to determine the structure of the approximated function. This led to a whole new field within \ac{ML}, \ac{xAI}, which tries to create methods that allow human users to understand the predictions made by an \ac{ML} model \citep{Vilone2021NotionsIntelligence}. Some popular examples include sensitivity analysis \citep{Zeiler2014VisualizingNetworks}, \ac{LIME} \citep{Ribeiro2016WhyClassifier}, and \ac{SHAP} values \citep{Lundberg2017APredictions}. This study uses the last method, as it is gaining in popularity and is now often used in ecology. For example, \cite{Masago2022EstimatingAlgorithms} use \ac{SHAP} values to investigate how inter-annual variation in the daily average temperature affected the first flowering date or the full blossom date of the Yoshino cherry trees in Japan. \cite{He2022ExplainableDistribution} construct a seagrass distribution model and explain the importance of environmental variables in the model and subsequent predictions. In \cite{Park2022InterpretationIntelligence}, an XGBoost model is trained to predict chlorophyll concentration, and they use \ac{SHAP} values to perform feature selection, as well as investigate feature importance. \ac{SHAP} values have a number of advantages over other methods for understanding the output of a model. First, \ac{SHAP} values are model-agnostic, which means that they can be used with any \ac{ML} model \citep{Lundberg2017APredictions}. Second, \ac{SHAP} values are able to account for interactions between features, which is something other methods are not able to do. Third, \ac{SHAP} values have an intuitive interpretation, which means that they are easy to understand and explain to others. Finally, \ac{SHAP} values have some desirable mathematical properties, such as local accuracy, missingness, and consistency \citep{AAS2021103502}.

An earlier study was conducted by \cite{Leblans2017PhenologicalWarming} at the same research sites in  Iceland \citep{Sigurdsson2016GeothermalStudy}, focusing on the phenology of subarctic grasslands. They used a short-term temporal dataset from 2013 to 2015 with curve function fitting analyses based on the methodology proposed by \cite{Zhang2003MonitoringMODIS} to determine seasonal (intra-annual) parameters (e.g. SOS). They found that the response towards earlier SOS in the warmed subarctic grasslands did not saturate at higher soil warming levels (\textit{i.e.,} +10°C). Therefore they concluded that growing seasons at high-latitudes grasslands are likely to continue lengthening with future warming. However, there was still quite a large unexplained inter-annual variability in their 3-year dataset, that warranted a further study \citep{Leblans2017PhenologicalWarming}. In the present study, we used six years of data instead of three years used by \cite{Leblans2017PhenologicalWarming}, which enabled us to look more deeply into inter-annual variability of \ac{NDVI} phenology and annual maximum values. The variables used for \ac{NDVI} phenology were the annual day numbers of \ac{SOS}, \ac{POS}, and \ac{PEAK} in each plot. The main aim was first to reanalyze the soil warming effect with conventional linear statistics as was done by  \cite{Leblans2017PhenologicalWarming},  and evaluate if those relationships held for a longer period. Secondly, we used \ac{ML} algorithms to explain the additional drivers impacting the unexplained inter-annual variation in the studied variables. The variables added in this step were air temperature, precipitation, and irradiation. As \ac{ML} methods are often not intuitive, we applied \ac{xAI} methods to gain further insights into the models.

Our objective was to study the relationship between soil temperature and vegetation phenology. More specifically, we studied this relationship using three vegetation phenology characteristics: \ac{SOS}, \ac{POS} and \ac{PEAK}. Additionally, we investigated the effect of other meteorological variables on these characteristics. 
To this end, we postulated following hypotheses:
\begin{enumerate}[label=\Alph*]
    \item \textbf{Soil warming} 
        \begin{enumerate}[label=\roman*.] 
            \item A higher soil temperature will introduce significantly earlier \ac{SOS}, as was found by \cite{Leblans2017PhenologicalWarming} for individual years.
            \item The \ac{POS} will take place at a similar time each year, regardless of the soil temperature. Plants must use some external trigger to ``know'' when to start to slow down growth and prepare for autumn. The prevailing theory suggests that for most plants, this is triggered by the length of the night, mediated through the phytochrome system \citep{Sigurdsson2001ElevatedStudy}.  
            This parameter has not been studied before.
            \item The \ac{PEAK} value will not be significantly related to soil temperature, as \cite{Verbrigghe2022SoilTopsoil} showed that there was no difference in above-ground biomass between the warming treatments.
        \end{enumerate}
    \item \textbf{Other meteorological variables} \\ We expect that \ac{ML} can identify other important controls for the previously observed inter-annual variability of \ac{NDVI} phenology and \ac{PEAK} values. Additionally, we expect that \ac{ML} can identify the importance of meteorological variables compared to the soil temperature. Out of the three additional meteorological variables, we hypothesized for both phenology and \ac{PEAK} values:
        \begin{enumerate}[label=\roman*.] 
            \item  Larger impact of meteorological variables compared to the soil temperature \citep{Xie2021SpringAlps}, as they also can impact the soil temperature \citep{Beer2018EffectsRegions, Tan2022InvestigatingChina}.            
            \item Within the meteorological variables air temperature's influence is expected to be the smallest due to its regulation of soil temperature \citep{Sigurdsson2016GeothermalStudy}, while precipitation may have an intermediate effect given consistently high soil water content in these areas \citep{Sigurdsson2016GeothermalStudy}. Additionally, a substantial impact of irradiance is hypothesized, particularly in consistently cloudy sub-Arctic climates \citep{Hou2015InterannualChina}.
        \end{enumerate}
\end{enumerate}

Ultimately, the contributions of this research advance our understanding of the relationships between soil temperature, other meteorological variables, and vegetation phenology. We achieve this goal by employing a methodology that exceeds standard practice, using \ac{ML} and \ac{SHAP} values.

\section{Materials and Methods}

\subsection{Data}

The study was carried out in the south of Iceland near the village of Hveragerdi on the ForHot site \citep{Sigurdsson2016GeothermalStudy}. Following an earthquake in May 2008, the bedrock of one unmanaged (cold) grassland field site underwent a disruption, resulting in the creation of areas with differently warmed soils. Another nearby grassland field site had had such warmed soil gradients for at least six decades, and those were not disturbed by the earthquake in 2008 \citep{Sigurdsson2016GeothermalStudy}. In spring 2013, five transects were selected in each field site, each with five permanent plots across the natural soil temperature gradients, resulting in a total of 50 studied plots. We categorized the plots according to their annual soil temperature range, as indicated in \cref{tab:fa:plotcategories}.
\begin{table}[H]
\renewcommand{\arraystretch}{1.25}
    \centering

    \caption{Category of the temperature range of the plots.} 
    {
    \begin{tabular}{p{0.27\textwidth}p{0.27\textwidth}p{0.11\textwidth}p{0.11\textwidth}p{0.11\textwidth}}
    \toprule
     Category & Temperature Range\\ 
    \midrule
    \makecell[l]{A}&\makecell[l]{Ambient}\\  \arrayrulecolor{lightgray}\midrule
    \makecell[l]{B}&+0.5 to 1°C \\ \midrule
    \makecell[l]{C}&\makecell[l]{+2 to 3°C}\\  \midrule
    \makecell[l]{D}&+3 to 5°C  \\ \midrule
    E &\makecell[l]{+5 to 10°C} \\ 
    \arrayrulecolor{defaultcolor}
    \bottomrule
    \end{tabular}
    }
     \renewcommand{\arraystretch}{1.0}
    \label{tab:fa:plotcategories}
\end{table}

\subsubsection{NDVI data}
\label{subsubsec:fa:ndvidata}
\begin{figure}[!tb]
    \centering

    \includegraphics[width=\linewidth]{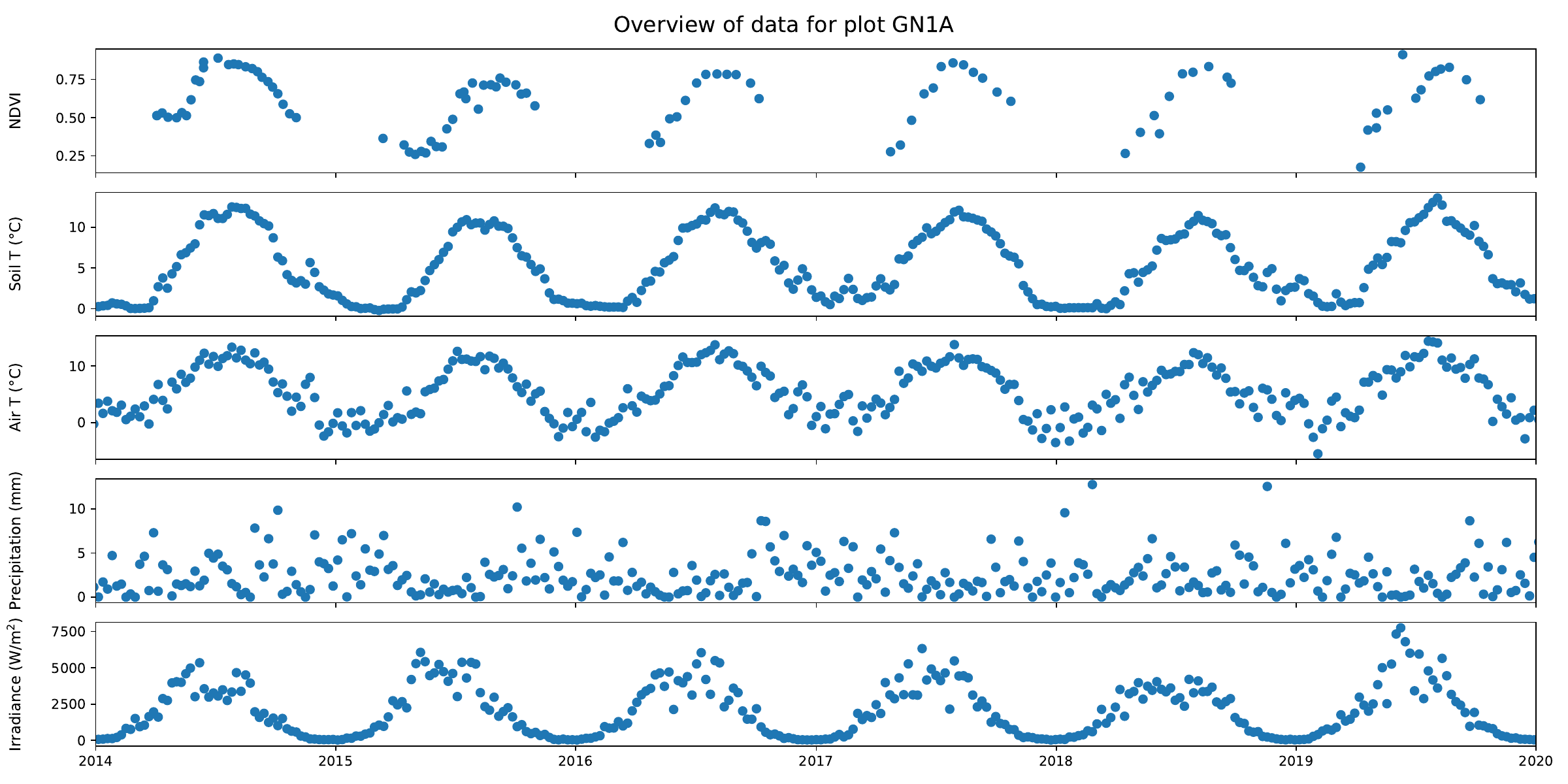}
    \caption{Overview of all available variables for plot GN1A (unwarmed control plot). Whereas the \ac{NDVI} and soil temperature (upper two figures) are unique for all 50 plots, the meteorological variables (bottom three figures) are the same for every plot.}
        \label{fig:fa:NDVI_data}

\end{figure}

The \ac{NDVI} was measured using a handheld instrument from SKYE Instruments (SpectraoSense2). From 2014 to 2019,  \ac{NDVI} measurements were done approximately bi-weekly from April to November, except during periods with continuous snow cover in early spring or late autumn. The measurements were always conducted on a clear day. We refer to \cite{Leblans2017PhenologicalWarming} for further information about the NDVI measurements. As can be seen in \cref{fig:fa:NDVI_data},  the \ac{NDVI} data clearly showed a seasonal pattern, with a higher \ac{NDVI} in the summer months. 

\subsubsection{Soil Temperature data}
\label{subsubsec:fa:soiltempdata}
The soil temperature at  a depth of 10 cm was monitored in all the permanent plots using HOBO TidbiT v2 Water Temperature Data Loggers (Onset Computer Corporation, USA) since the spring of 2013  \citep{Sigurdsson2016GeothermalStudy}. In \cref{tab:fa:plotcategories}, the different soil warming categories with their accompanying temperature range are given, while  \cref{fig:fa:NDVI_data} shows the data for one of the 50 plots used in this study. The main soil warming effect was an approximately constant shift in temperature across the seasons, as shown by \cite{Sigurdsson2016GeothermalStudy}.

\subsubsection{Meteorological data}
\label{subsubsec:fa:meteodata}
Meteorological variables for the period 2014-2019, including irradiance (global radiation), precipitation, and air temperature, were obtained from a weather station in Reykjavík, about 40 km from the research site (data courtesy of the Icelandic Meteorological Institute). This is the closest station where irradiance is measured. We aggregated the data by taking the average on a weekly resolution scale. 
As the nearest weather station is not located precisely at the plots, we rely on this data as a proxy for the actual weather conditions at the ForHot site. We therefore also assume that the weather conditions are the same for all plots during each year. In \cref{fig:fa:NDVI_data}, the three bottom panes show all meteorological variables measured in the relevant period.

\subsection{Data analysis}
\subsubsection{Estimating the NDVI seasonal characteristics}
\label{doublelogistic}

In order to estimate the intra-annual characteristics in each permanent plot during each growing season, a double logistic curve was used, based on the approach of \cite{Zhang2003MonitoringMODIS}. We require that the two logistic curves transition into each other continuously, such that the resulting function is differentiable at every point. These requirements result in the following formula for the estimated \ac{NDVI}:

\begin{equation}
\widehat{NDVI}(x) =  \begin{dcases} 
      \frac{c}{1 + e^{b_1 \cdot (x - a_1)}} + d & x \leq p \\
      -\frac{c}{1 + e^{b_2 \cdot (x - a_2)}} + d + c & x > p
   \end{dcases}
\end{equation}

where the parameters $a_1$, $a_2$, $b_1$, $b_2$, $c$, $d$ and $p$ are fitted to a season's \ac{NDVI} data and $x$ represents the week number $([0, 52])$ of the year. The parameter $p$ has an important interpretation, as it is defined as the date of the peak of the season, i.e., where the maximal \ac{NDVI} value is reached.

The best fit for the curve parameters is found using the Trust Region Reflective algorithm \citep{Conn2000TrustMethods}. This generally robust optimization method minimizes the \ac{MSE} between the predicted \ac{NDVI} curve and the \ac{NDVI} data points. After the curve parameters have been fitted, we extracted the start \ac{SOS}, \ac{POS} and \ac{PEAK} for each plot in each year.

\begin{figure}[H]
    \centering

    \includegraphics[width=0.6\linewidth]{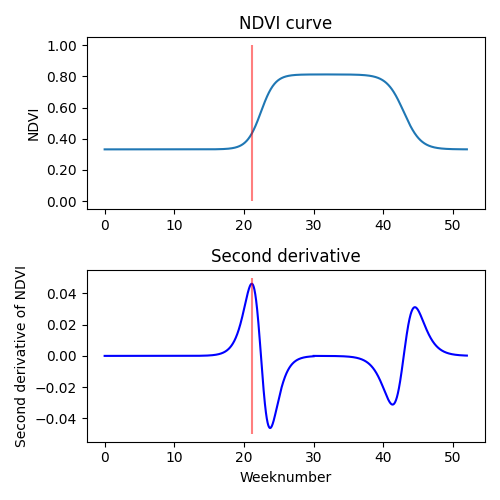}
    \caption{The \ac{SOS} is estimated based on the second derivative of the fitted \ac{NDVI} curve. The \ac{SOS} is defined as the week when the \ac{NDVI} curvature increases the most, and is indicated with a red line.}         
        \label{fig:fa:seasoncharacteristics}

\end{figure}

The second derivative of the fitted curves is used to calculate \ac{SOS}. This estimated parameter is considered to be the time of year when the \ac{NDVI} curvature increases the most. As shown in \cref{fig:fa:seasoncharacteristics} 
, the estimated start of season is the moment in time when the second derivative of the first logistic function is maximal: 

\begin{align}
 \widehat{SOS} &= \argmax_x -\frac{cb_1^2e^{b_1\left(x-a_1\right)}\left(-e^{b_1\left(x-a_1\right)}+1\right)}{\left(1+e^{b_1\left(x-a_1\right)}\right)^3} \\
\widehat{POS} &= p\\
\widehat{PEAK} &= \widehat{NDVI(p)}
\end{align}

where $\widehat{SOS}$ indicates the estimated start of the season, $\widehat{POS}$ the date of the peak of the season, and $\widehat{PEAK}$ the maximum value of the \ac{NDVI}.

\subsection{Statistical modeling and machine learning}

\subsubsection{Linear regression}
\label{subsec:fa:linreg}
After having obtained the start and peak season for each plot and year, we fitted a linear regression model to the \ac{SOS}, \ac{POS}, and \ac{PEAK} with the average soil temperature in each plot and year as the independent variable. Linear regression models were fitted using the implementation of ordinary least squares in statsmodels 0.13.2 \citep{Seabold2010Statsmodels:Python} in Python 3.9.13. This implementation also allowed us to calculate the p-values from a t-test of the slope and intercept of the linear model. 

\subsubsection{Machine learning}
\label{subsec:fa:machine_learning}
To further study the inter-annual variability in the above responses, we trained \ac{MLP}s including the meteorological variables, one to predict the start of the season, one to predict the peak of the season, and one to predict the height of the peak of the season. These \ac{MLP}s have a total of 79 input variables (or input nodes). These input variables are the average weekly air temperature in the first 26 weeks of the year, the average weekly precipitation in the first 26 weeks of the year, the average weekly radiation in the first 26 weeks of the year, and the average soil temperature during a year.
The \ac{MLP}s   were implemented using the MLPRegressor class in scikit-learn 1.1.3 \citep{Pedregosa2011Scikit-learn:Python}. We optimized the hyperparameters of the \ac{MLP}s   separately for the three target variables using a 5-fold \ac{CV} grid search implemented by Optuna 3.1.0 \citep{Akiba2019Optuna:Framework}. A description of the hyperparameters, their explored ranges, and optimal values can be found in \cref{tab:fa:gridsearch_results}. We used the \ac{MSE}, \ac{MAE} and $r^2$ to assess the performance of the models. In the grid search, only the \ac{MSE} was used to find the optimal combination of hyperparameters. Prior to conducting the grid search, we split the data in a train and test set, containing respectively 80\% and 20\% of all samples.

\begin{table}[H]
\renewcommand{\arraystretch}{1.25} 
    \centering
    \caption{Overview of the explored ranges of hyperparameters used in the Optuna grid search. The optimal values for the three different regression tasks are displayed in the right-most three columns.} 
    {
    \begin{tabular}{p{0.27\textwidth}p{0.27\textwidth}p{0.11\textwidth}p{0.11\textwidth}p{0.11\textwidth}}
    \toprule
     Description & Range & \ac{SOS} & \ac{POS} & \ac{PEAK}\\ 
    \midrule
    \makecell[l]{Number of neurons\\ in first layer}&\makecell[l]{int: 10, 20, \dots,  100}&100&70&30 \\  \arrayrulecolor{lightgray}\midrule
    \makecell[l]{Number of neurons\\ in second layer}&int: 0, 10, \dots,  100&0&0&100 \\ \midrule
    \makecell[l]{Strength of the L2\\ regularization term}&\makecell[l]{float: 1e-4 --- 1e-1\\ logscale}&0.0290&0.0010&0.0606\\  \midrule
    \makecell[l]{the solver for\\ weight optimization}&adam, lbfgs&adam&adam&adam \\ \midrule
    initial learning rate&\makecell[l]{float: 1e-4 --- 1e-1\\ logscale}&0.0031&0.0003&0.0028 \\ \midrule
    \makecell[l]{learning rate\\ schedule for\\ weight updates}&constant, adaptive&constant&adaptive&adaptive \\ \midrule
    \makecell[l]{maximum number\\ of iterations}&\makecell[l]{int: 1000, 2000, \dots,\\ 10000}&8000&8000&8000 \\ \midrule
    \makecell[l]{maximum number\\ of iterations with no\\ improvement}&int, 10, 20, \dots, 100&20&50&100 \\
    \arrayrulecolor{defaultcolor}
    \bottomrule
    \end{tabular}
    }
     \renewcommand{\arraystretch}{1.0}
    \label{tab:fa:gridsearch_results}
\end{table}

\subsubsection{SHAP values}   
The 79 input features are not equally important, and will influence the predictions in different ways. \ac{SHAP} values are a way of understanding which features of a data set are the most important to predict the output of a \ac{ML} model. They are calculated by taking into account how the model output would change if each feature were to be turned on or off. In this way, the \ac{SHAP} values assign an importance score to each feature. Every prediction made by the model can be decomposed by the \ac{SHAP} values for every feature, as the sum of the \ac{SHAP} values equals the model output. 

After training the \ac{MLP} models, we compute \ac{SHAP} values using the model-agnostic Kernel \ac{SHAP} method to understand the learned model and which features are most important in predicting the start and (height of the) peak of the greening season. We used the implementation in the Python \ac{SHAP} package (0.41.0) \citep{Lundberg2017APredictions}.

\section{Results}
\subsection{The logistic fitting}
For most plots and years, good fits were found for the double logistic curves that were fitted to the intra-annual individual plot \ac{NDVI} data, with an average $r^2$ of 0.942 ($\pm$ 0.095). However, for 5.8\% of all plots and years, the data did not follow a double sigmoid curve, and the $r^2$ value was lower than 0.80. These curves were not included in the analysis.  The mean estimated \ac{SOS} was week 20.41 ($\pm$ 2.40), the mean estimated \ac{POS} was week 29.97 ($\pm$ 3.27), and the mean estimated \ac{PEAK} was 0.842 ($\pm$ 0.071) across all the soil warming treatments.

\subsection{The average response to soil temperature}

\begin{figure}[H]
    \centering

    \includegraphics[width=\linewidth]{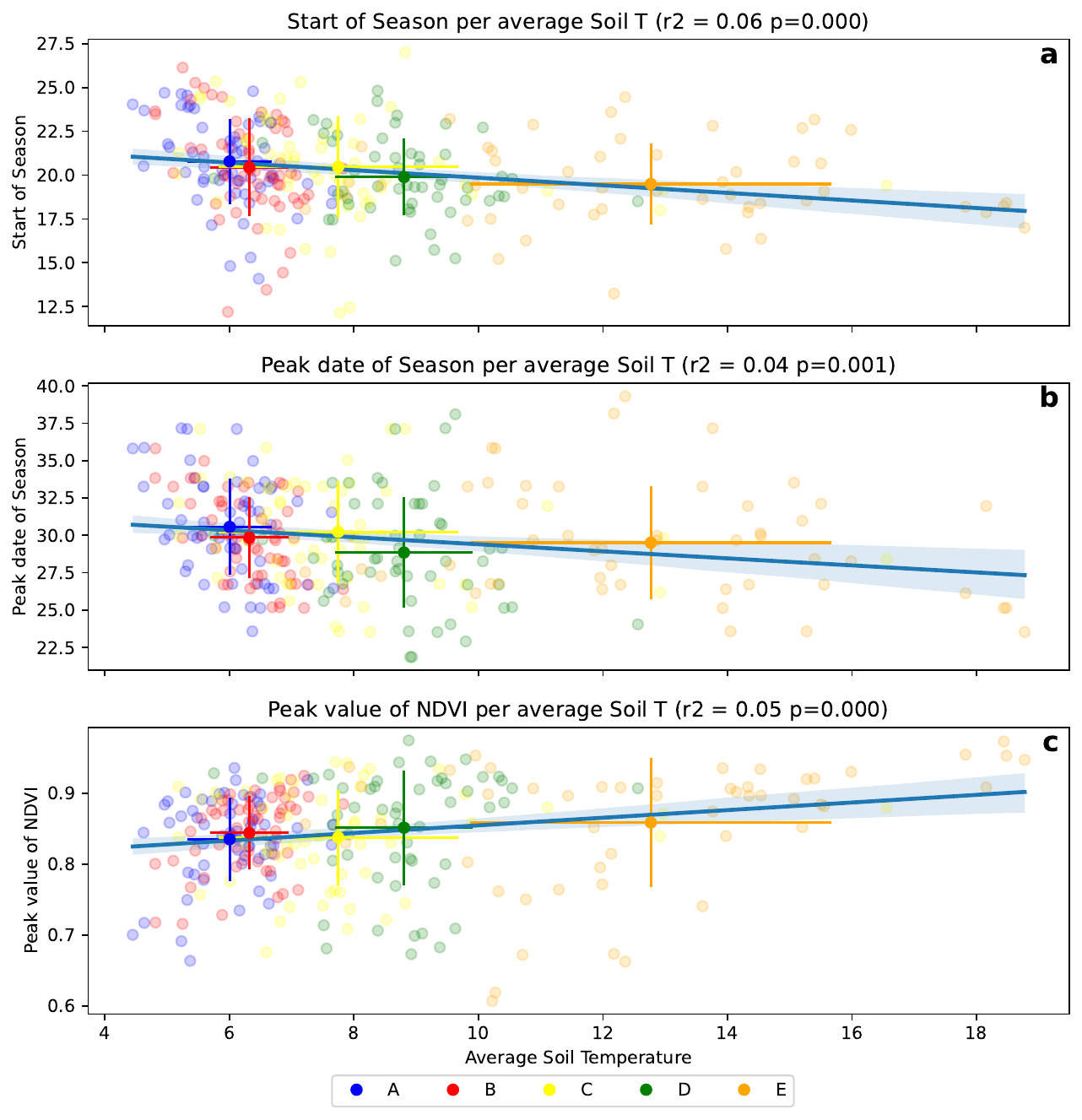}
    \caption{Linear model that predicts the start of the season (a),  the peak date of the season (b) and the peak value of \ac{NDVI} (c), based on the average annual soil temperature. The color indicates the soil warming category where the blue points are A plots, the red points are B plots, the yellow points are C plots, the green points are D plots, and the orange points are E plots. All models had a significant relationship between the average soil temperature and the studied \ac{NDVI} curve parameter. (See \cref{tab:fa:linregequations}) }
        \label{fig:fa:linearmodels}

\end{figure}

\cref{fig:fa:linearmodels} shows the linear relationship found between the average annual soil temperature and the three \ac{NDVI} characteristics found by the double-logistic curves. The parameters of the linear model are given in \cref{tab:fa:linregequations}. A significant linear relationship was found between average soil temperature and \ac{SOS} ($p < 0.001$), \ac{POS} ($p = 0.001)$ and \ac{PEAK}  \ac{NDVI} ($p < 0.001$) (\cref{fig:fa:linearmodels,tab:fa:linregequations}). The relationship between soil temperature and \ac{SOS} was negative, with an estimated coefficient of -0.2160 ($\pm$ 0.053). This means that for every 4.63 degrees of soil warming, the greening season starts a week earlier. Otherwise stated, the \ac{SOS} happens 1.52 days earlier per degree of soil warming when derived across multiple years. Similarly, we see that the date of the \ac{NDVI} peak shifted forward. The estimated coefficient of -0.2353 ($\pm$ 0.07) indicates that for every 4.25 degrees of soil warming, the \ac{NDVI} peaks a week earlier, or the \ac{POS} occurs 1.65 days earlier per degree of soil warming. 
Finally, the \ac{PEAK} value of the \ac{NDVI} curve increased slightly with increasing soil temperature. 

Although the linear relationships that were observed between average soil temperature and \ac{SOS}, \ac{POS}, and \ac{PEAK} were significant (\cref{fig:fa:linearmodels}), we also observed a lot of unexplained variance, which is indicated by the relatively low $r^2$ values in \cref{tab:fa:linregequations}.

\begin{table}[H]
    \centering
    \caption{The parameters describing the results of the linear models, where different variables are fitted against the average soil temperature over a whole year. The \ac{SOS} and \ac{POS} are measured in weeks, while the intercept is measured in degrees Celsius.} 
    \begin{tabular}{*5c}
    \toprule
    Target variable & Slope & Intercept & r$^2$&p-value\\
    \midrule
    \ac{SOS} & $-0.216 \pm 0.052$ & $22.011 \pm 0.454$ &0.06&0.000 \\
    \ac{POS} & $-0.235 \pm 0.070$& $31.755 \pm 0.607$ &0.04&0.001\\
    \ac{PEAK} & $0.005 \pm 0.001$& $0.801 \pm 0.013$ &0.05&0.000 \\
    \bottomrule
    \end{tabular}
    \label{tab:fa:linregequations}
\end{table}

\subsection{The machine learning approach}

To explain a larger part of the variance, the possibility of predicting  characteristics of the \ac{NDVI} curve using \acp{MLP}, based on both the soil temperature and meteorological variables, was investigated. The performance of the \acp{MLP} can be found in \cref{tab:fa:ML_results}. From \cref{tab:fa:linregequations,tab:fa:ML_results}, it becomes evident that the inclusion of the meteorological variables and the utilization of \acp{MLP} enabled us to explain a significantly larger part of the variance compared to the linear models.

\begin{table}[H]
    \centering
    \caption{Model performance of \ac{MLP} after a 5-fold cross validation grid search. The test set consists of 20\% of the total data, and is split evenly across the years of data taking. The naive \ac{MSE} (\ac{MAE}) is the \ac{MSE} (\ac{MAE}) when the mean of all training samples is used as the prediction.}
    \begin{tabular}{*5c}
    \toprule
         Target & 5-fold \ac{CV} \ac{MSE}& Test \ac{MSE} (naive) & Test \ac{MAE} (naive) & Test r$^2$\\ 
         \midrule
         \ac{SOS}&3.408&4.760 (7.102)&1.521 (2.095)&0.322\\
         \ac{POS}&7.933&8.943 (11.103)&2.473 (2.696)&0.192\\
         \ac{PEAK}&0.004&0.004 (0.006)&0.053 (0.063)&0.248 \\
     \bottomrule
    \end{tabular}
    \label{tab:fa:ML_results}
\end{table}

\begin{figure}[H]
    \centering

    \includegraphics[width=\linewidth]{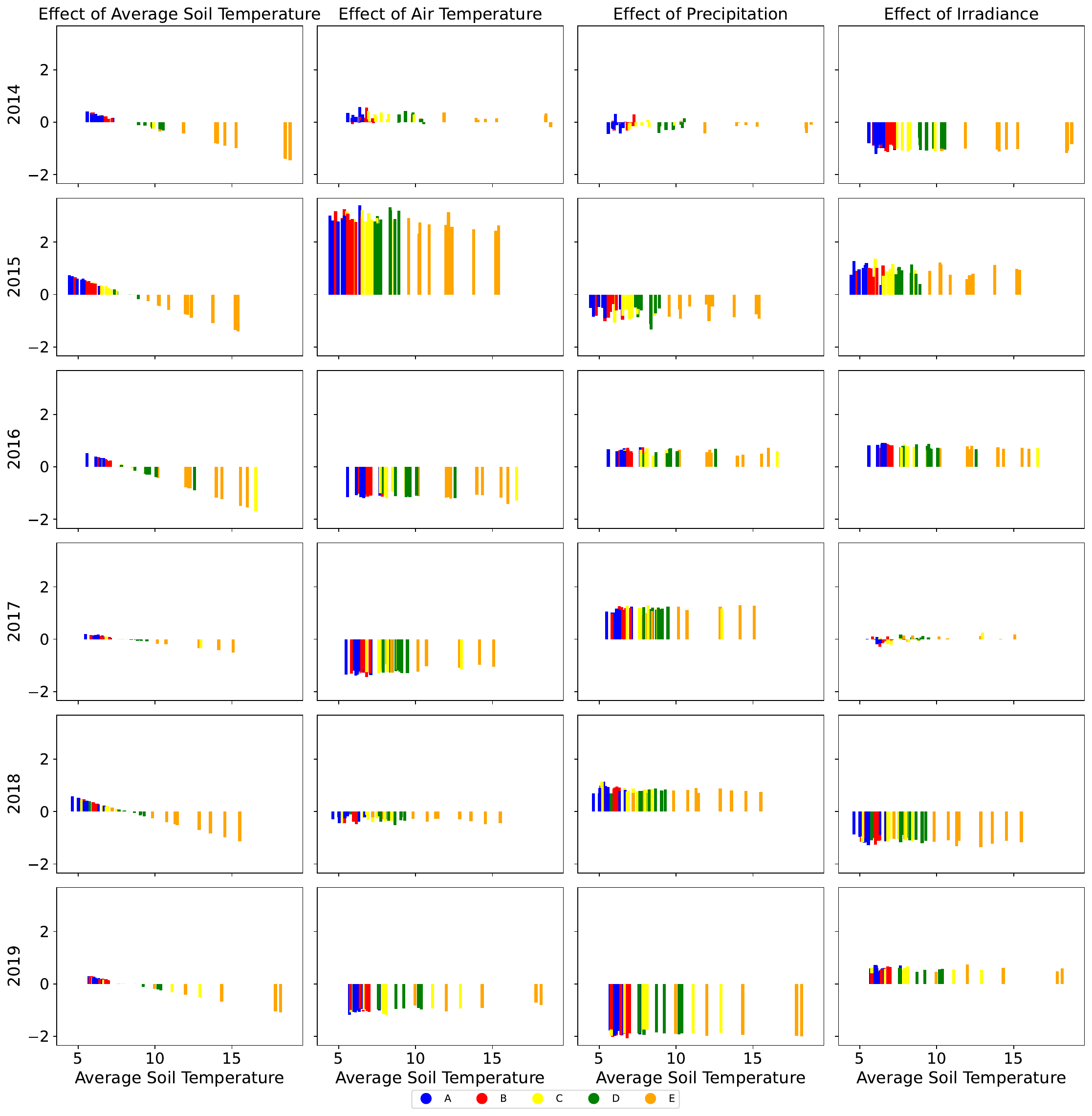}
    \caption{\ac{SHAP} values of multi-layer perceptron that predicts the start of the greening season based on the average soil temperature, air temperature, precipitation, and radiation. The color indicates the soil warming category where the blue bars are A plots, the red bars are B plots, the yellow bars are C plots, the green bars are D plots, and the orange bars are E plots.}
        \label{fig:fa:SHAP_SOS}

\end{figure}

\begin{figure}[H]
 
    \centering
    \includegraphics[width=\linewidth]{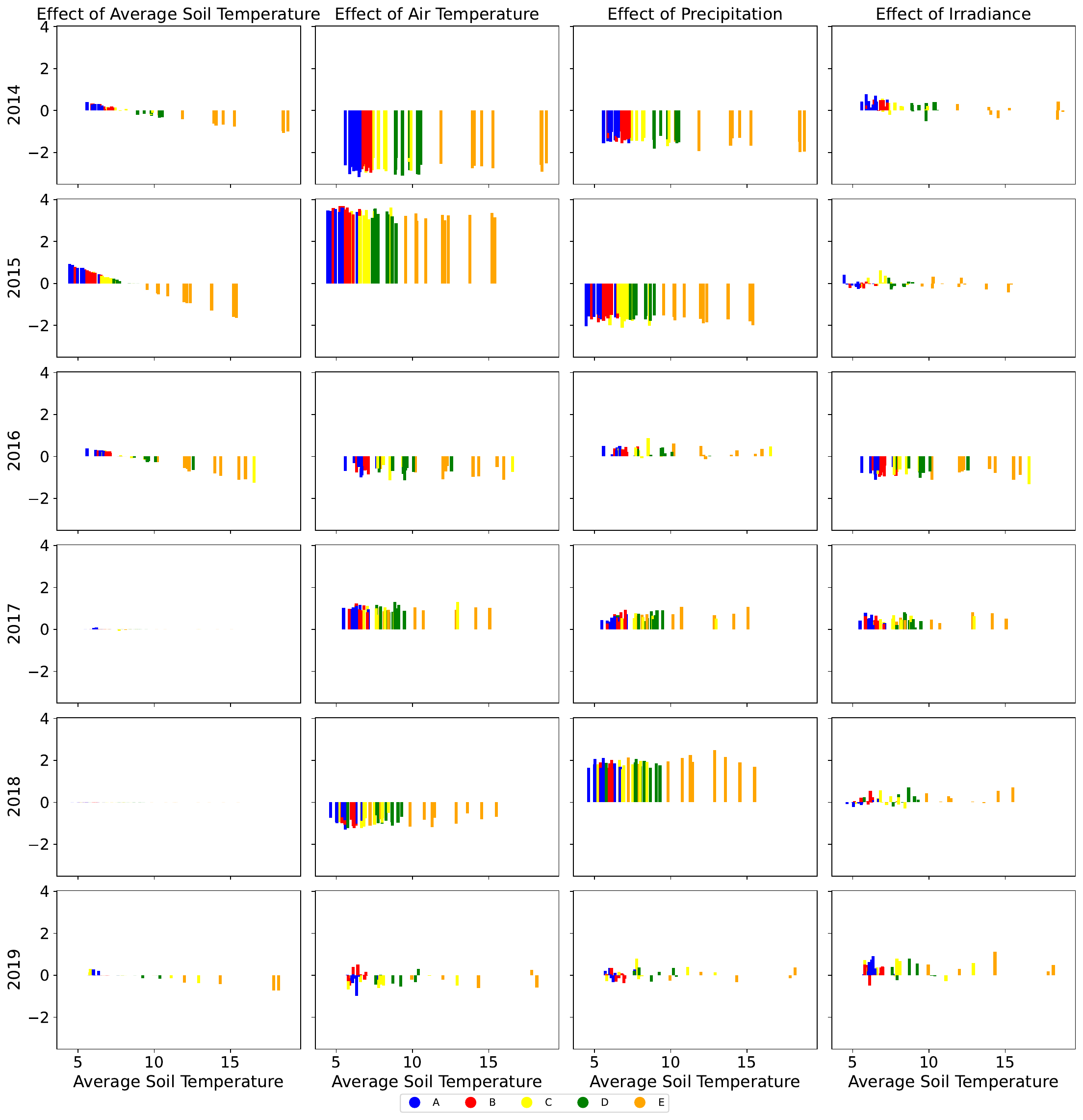}
    \caption{\ac{SHAP} values of multi-layer perceptron that predicts the peak of the greening season (POS)  based on the average soil temperature, air temperature, precipitation, and radiation. The color indicates the soil warming category where the blue bars are A plots, the red bars are B plots, the yellow bars are C plots, the green bars are D plots, and the orange bars are E plots.}
        \label{fig:fa:SHAP_POS}

\end{figure}

\begin{figure}[H]
    \centering
    \includegraphics[width=\linewidth]{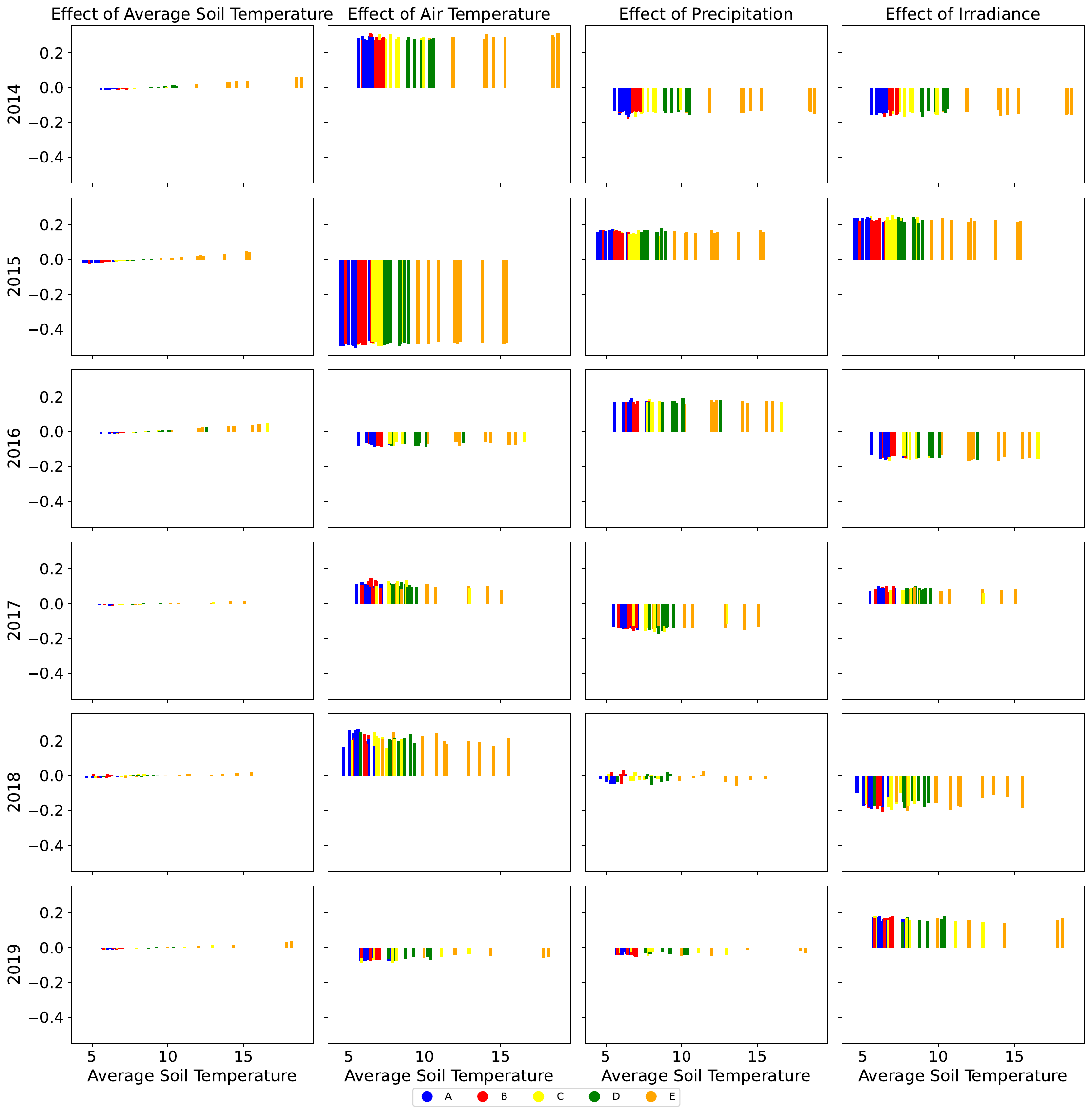}
    \caption{\ac{SHAP} values of multi-layer perceptron that predicts the peak \ac{NDVI}  based on the average soil temperature, air temperature, precipitation, and radiation. The color indicates the soil warming category where the blue bars are A plots, the red bars are B plots, the yellow bars are C plots, the green bars are D plots, and the orange bars are E plots.}
        \label{fig:fa:SHAP_PEAK}

\end{figure}

\begin{figure}[H]
 \centering
     \begin{subfigure}[b]{0.325\textwidth}
         \centering
         \includegraphics[width=\textwidth]{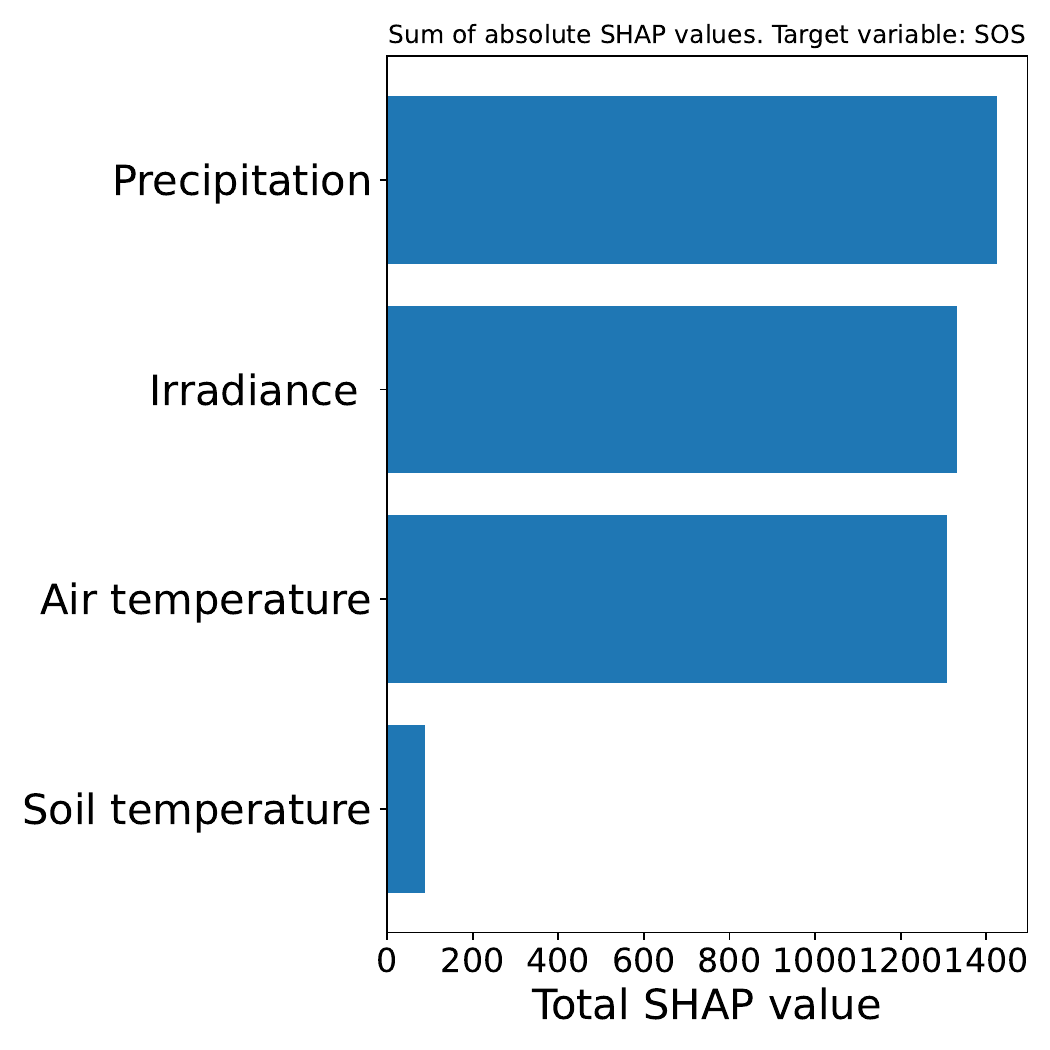}
         \caption{Start of season}
         \label{fig:fa:SHAP_tot_SOS}
     \end{subfigure}
     \hfill
     \begin{subfigure}[b]{0.325\textwidth}
         \centering
         \includegraphics[width=\textwidth]{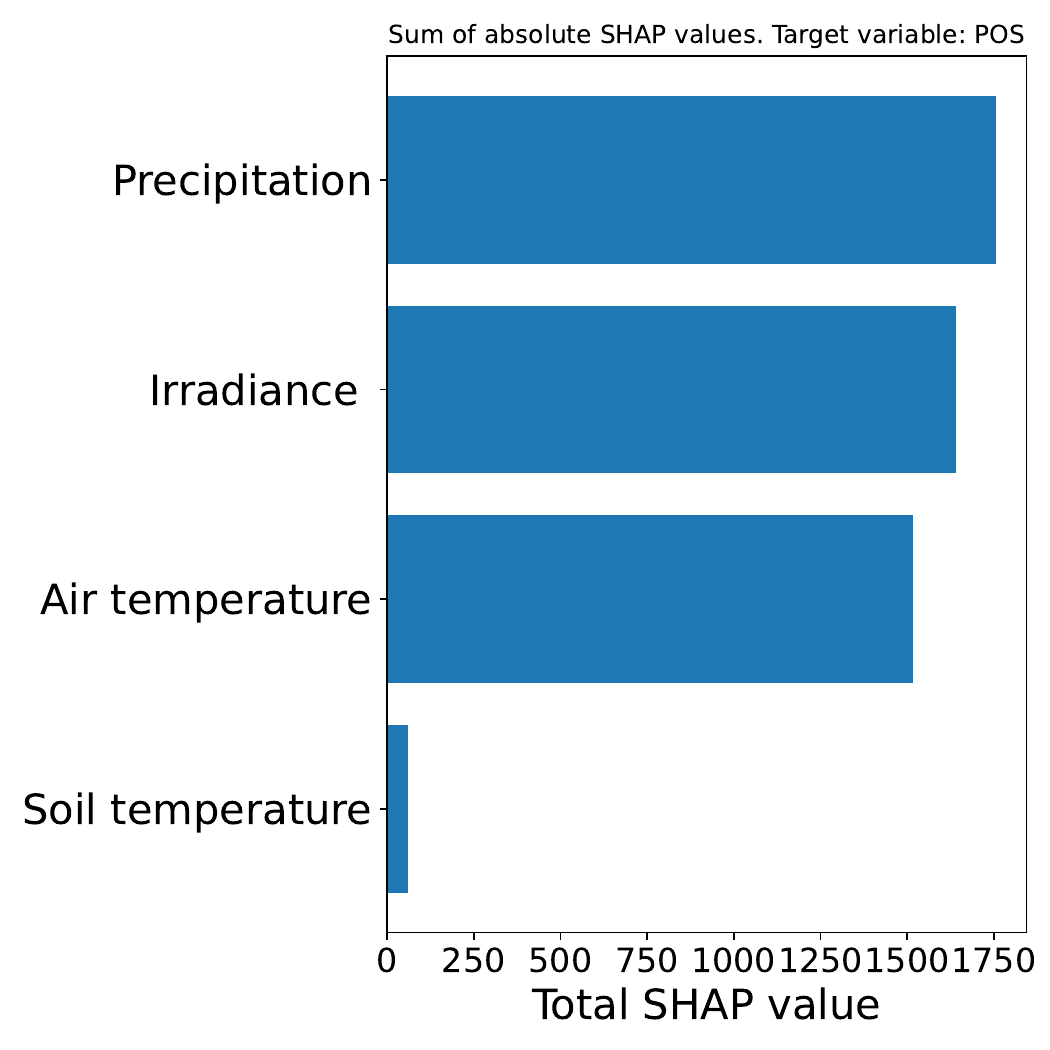}
         \caption{Peak of season}
         \label{fig:fa:SHAP_tot_POS}
     \end{subfigure}
     \hfill
     \begin{subfigure}[b]{0.325\textwidth}
         \centering
         \includegraphics[width=\textwidth]{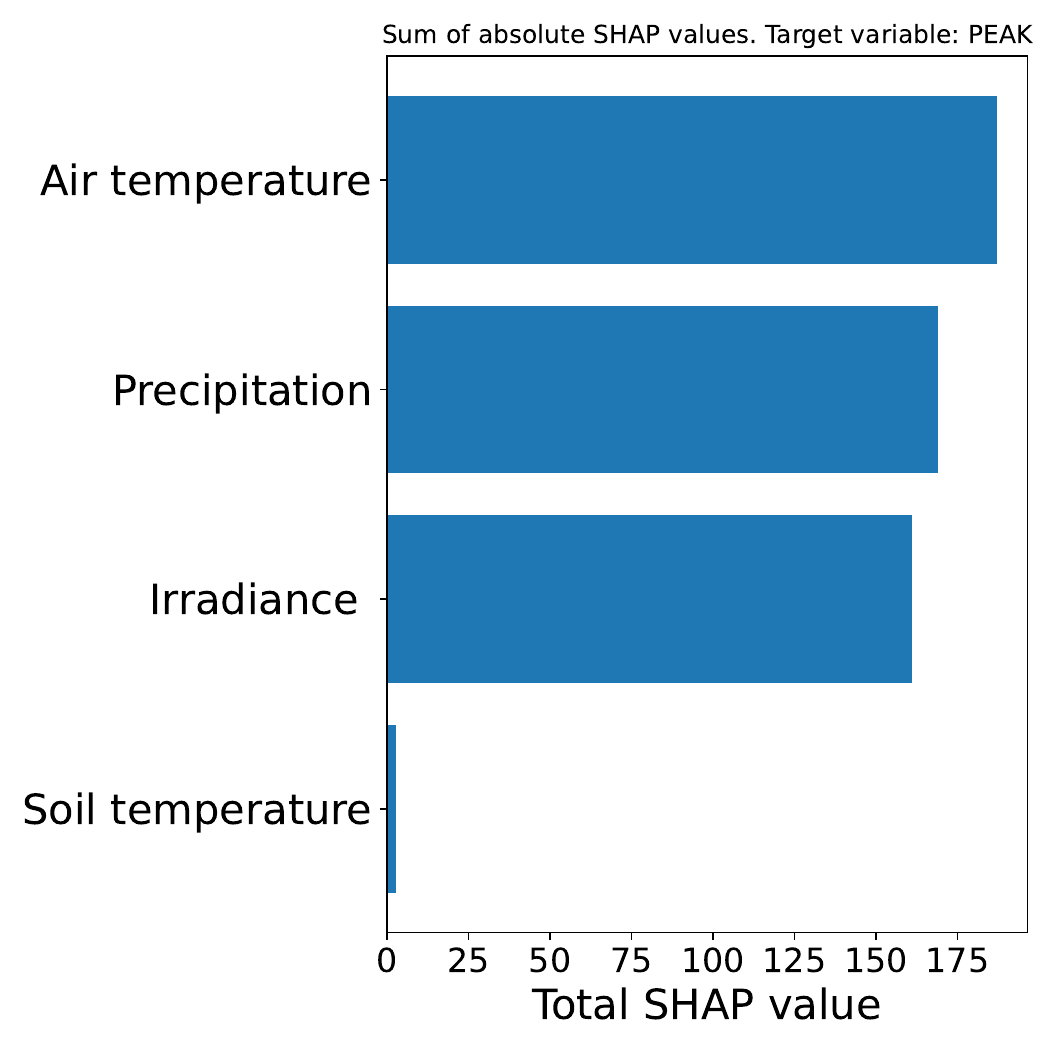}
         \caption{Peak NDVI value}
         \label{fig:fa:SHAP_tot_PEAK}
     \end{subfigure}
        \caption{Sum of the absolute SHAP values as defined in \cref{eq:fa:SHAP_absolute_sum}.}
        \label{fig:fa:SHAP_totals}
\end{figure}

To investigate the impact of a given feature on the predictions made by the model, we calculated \ac{SHAP} values for all three \ac{MLP}s. These can be found in 
\cref{fig:fa:SHAP_SOS}, \cref{fig:fa:SHAP_POS} and \cref{fig:fa:SHAP_PEAK} 
for the \ac{SOS}, \ac{POS} and \ac{PEAK}, respectively. In these figures, we separate the six years to investigate the annual variation in the \ac{SHAP} values. To obtain the \ac{SHAP} value for one meteorological variable, we summed the \ac{SHAP} values of the 26 weekly averages, as shown in \cref{eq:fa:SHAP_meteorological}. Next, we calculated the sum of absolute values of the \ac{SHAP} values \mbox{A\_SHAP} for the four remaining features for all $n$ samples, as shown in \cref{eq:fa:SHAP_absolute_sum}. By taking the absolute value and adding it over all years, we can investigate the total impact of a feature on the prediction, regardless of the direction of the impact. The results for the \mbox{(A\_SHAP)} values are shown in \cref{fig:fa:SHAP_totals}.

\begin{align}
    \text{SHAP}_{feature} &= \sum_{week=1}^{26} \text{SHAP}_{feature, week} \label{eq:fa:SHAP_meteorological}\\
    \text{A\_SHAP}_{feature} &= \sum_{i}^n |\text{SHAP}_{feature, i}| \label{eq:fa:SHAP_absolute_sum}
\end{align}

When interpreting \cref{fig:fa:SHAP_SOS,fig:fa:SHAP_tot_SOS}, we see that the meteorological variables had the largest impact on the prediction of the \ac{SOS}. However, within each year, this impact was approximately constant. The intra-annual variation in the \ac{SOS} was clearly the result of soil warming. In fact, the Pearson correlation between soil temperature and its accompanying \ac{SHAP} values was -0.93, meaning that the higher the soil warming, the earlier the season started each year. All Pearson correlation values can be found in \cref{tab:fa:pearson_SHAP}.

From \cref{fig:fa:SHAP_tot_POS,fig:fa:SHAP_tot_PEAK}, we can also conclude that the three meteorological variables also had the largest impact on the predictions of the \ac{POS} and \ac{PEAK}. From \cref{tab:fa:pearson_SHAP}, we can see that the \ac{POS} was earlier and the \ac{PEAK} value of the \ac{NDVI} was higher with increasing soil temperature, as they had a Pearson correlation coefficient of -0.85 and 0.91, respectively. For the \ac{POS}, \cref{fig:fa:SHAP_POS} indicates that the size and direction of the \ac{SHAP} effect for the three meteorological variables shifts significantly over the years, while the smaller effect of the soil temperature is relatively stable across the six years and drives the intra-annual variation within the dataset.

\begin{table}[H]
    \centering
    \caption{Pearson correlation coefficient between the average soil temperature and its corresponding \ac{SHAP} values.}
    \begin{tabular}{*2c}
    \toprule
         Target variable & Pearson correlation\\
         \midrule
  
         \ac{SOS}&-0.93\\
         \ac{POS}&-0.85\\
         \ac{PEAK}&0.91\\

     \bottomrule
    \end{tabular}

    \label{tab:fa:pearson_SHAP}
\end{table}

\section{Discussion}

The purpose of this study was to investigate the relationship between soil temperature and \ac{NDVI} using \ac{ML} techniques. The discussion will focus on 
emphasizing the novelties of this work, 
addressing the hypotheses presented in the paper, discussing the findings in relation to previous research, and highlighting the implications of the results.

\subsection{Using machine learning to study vegetation phenology}
Currently, the standard practice in vegetation phenology studies using \ac{NDVI} consists of using simple statistical methods such as linear regression \citep{Leblans2017PhenologicalWarming}. However, our results clearly indicate that, after applying linear regression, a large amount of unexplained variance remains. Our work goes one step further by building \ac{ML} models that incorporate meteorological variables and are capable of modeling nonlinear relationships. Using \acp{MLP}, we successfully managed to explain a larger part of the inter-annual variance. Additionally, \ac{SHAP} values allowed us to gain more insights into the decisions made by an otherwise black-box \ac{MLP}.

\subsection{Effect of the soil temperature on SOS, POS, and PEAK in subarctic grassland}
The first hypothesis stated that a higher soil temperature would lead to an earlier \ac{SOS} based on previous research by \cite{Leblans2017PhenologicalWarming}. Such responses have also been found when past changes in \ac{NDVI} have been related to changes in annual, seasonal or monthly temperatures \citep{Potter2020ChangesDecades, Arndt2019ArcticAlaska, Karlsen2014SpatialData}.   

The findings of this study supported this hypothesis, as a significant relationship was observed between average soil temperature and the start of the greening season. The negative coefficient (-0.2160) indicates that \ac{SOS} occurs 1.5 days earlier per degree of soil warming across the six years. This finding was consistent with a recent analysis from the International Tundra Experiment covering up to 20 years of data from 18 sites and 46 open-top chamber warming experiments across the Arctic, sub-Arctic, and alpine ecosystems. They observed a 0.73-day earlier start of the greening season, in an environment where the average air warming was 1.4 °C and the soil warming approximately half of that \citep{Collins2021}. Our finding was also consistent with previous research at the same ForHot site, as  \cite{Leblans2017PhenologicalWarming} found that on average, the \ac{SOS} occurred 1.6 days earlier for every degree of soil warming.

The second hypothesis stated that the date of the \ac{POS} would occur at a similar time each year, regardless of soil temperature. The results of this study provide new information by showing that the \ac{POS} was similarly advanced by higher the soil temperature as the \ac{SOS}, or approximately 1.7 days earlier per degree of soil warming. The hypothesis was therefore rejected. This finding suggests that, in our sub-Arctic grasslands, day length might not be the primary factor influencing the timing of the \ac{POS} \citep{Sigurdsson2001ElevatedStudy}.

The third hypothesis proposed that the \ac{PEAK} \ac{NDVI} would not be significantly related to soil temperature, based on previous research by \cite{Verbrigghe2022SoilTopsoil}, who had not found significant differences in vegetation biomass across the warming gradients. However, the findings of this study indicate a slight increase in the \ac{PEAK} value with increasing soil temperature. Although the relationship was not as strong as for the \ac{SOS} and \ac{POS}, it suggests that higher soil temperatures may contribute to higher NDVI peak values. It is worth noting that while \ac{NDVI} is often used to estimate vegetation biomass \citep{Zhang2016, Lumbierres2017, Perry2022}, it is not measuring it directly, but rather the amount of chlorophyll per surface area \citep{Huang2021ASensing}. Therefore, ``arctic greening'' measured using the \ac{NDVI}, could occur without any changes in vegetation biomass, if the plants are getting ``greener'' due to a higher nutrient content in warmer soils. Further research is needed to better understand this relationship and its underlying mechanisms.

\subsection{Effect of the other meteorological variables}
Hypothesis B focused on the impact of other meteorological variables (air temperature, precipitation, and irradiance) on the inter-annual variability of the NDVI phenology and \ac{PEAK} values, and the potential of \ac{ML} to identify their importance. The results of the \ac{ML} analysis using \acp{MLP} showed that these variables have a strong impact on the predictions of the \ac{SOS}, \ac{POS}, and \ac{PEAK}, and the $r^2$ values of the \acp{MLP} were much higher than those obtained by the linear regression.

The SHAP values also provided information on the relative importance of these variables. It was noteworthy that the three meteorological variables had a much larger impact on the predictions than the soil warming data. However, the intra-annual variation in the \ac{SOS}, \ac{POS}, and \ac{PEAK} was found to be influenced by the soil temperature.

Contrary to our initial hypothesis, the SHAP values did not indicate significant differences among the meteorological parameters, making it challenging to prioritize their impact as hypothesized. However, collectively, these meteorological factors exhibited a considerably higher influence on the predictions compared to the soil warming data. Therefore, our findings not only contribute to understanding the dominant impact of meteorological parameters on vegetation dynamics but also emphasize the need for continued research to explain the interdependencies and potential interactions between these factors. 

\subsection{Methodological considerations}
It is important to note some limitations of the study. The analysis focused on a specific location in Iceland, and the results may not be directly applicable to other regions. The study period also covered a limited period of time (2014-2019), and longer-term data would provide a more comprehensive understanding of the inter-annual variation in \ac{NDVI}.  Furthermore, the meteorological data does not have the same spatial resolution as the \ac{NDVI} or soil temperature data, as we assumed that the weather conditions were the same across all plots. 

The \ac{SHAP} values should also be interpreted with caution. Although they are model-agnostic, we can only draw valid conclusions if the model generalizes well. That is, if it has an acceptable test set performance \citep{Molnar2021GeneralModels}. Furthermore, the \ac{SHAP} values do not have a causal interpretation \citep{frye2020asymmetric}. We cannot assume that if the variable X has a large impact on the prediction of Y, then X causes Y. On the contrary, Y might cause X, X and Y could both be caused by a confounding variable, or they could have no causal relationship at all.

Nevertheless, this study produces valuable insights and provides clear directions for future research. Our promising results, achieved by applying \ac{ML} in a vegetation phenology study, emphasize the potential of this approach in advancing our understanding of seasonal plant characteristics based on \ac{NDVI} data. They can also be viewed as a starting point for other analyses in a broader ecological context.

In the future, it would be interesting to consider other model architectures or methodologies, for example, XGBoost \citep{Chen2016XGBoost:System}. Additionally, other \ac{xAI} approaches like \ac{LIME} \citep{Ribeiro2016WhyClassifier} could be considered, allowing comparison between different \ac{xAI} approaches.

\section{Conclusions}

Our results only partly supported our hypotheses regarding the effect of soil temperature on the timing of the \ac{SOS}, timing of the \ac{POS}, and peak \ac{NDVI} values. We observed a significant relationship between soil warming and the timing of \ac{SOS} and \ac{POS}, indicating that higher soil temperatures lead to an earlier onset of the growing season but also to a similar shift in the timing of the \ac{POS}. Moreover, the peak \ac{NDVI} values showed a slight increase with higher soil temperatures. Furthermore, we explored the impact of meteorological variables, including air temperature, precipitation, and irradiance, on vegetation phenology and its inter-annual variation. The use of \ac{SHAP} values allowed us to gain insight into the relative importance and contribution of each meteorological variable to the predictions. It became evident that the three meteorological variables had the largest impact on the prediction of \ac{SOS}, \ac{POS}, and \ac{PEAK} \ac{NDVI} values across the six years. However, within a given year, the impact of the three meteorological variables remained approximately equal, while the variations in phenological characteristics were primarily driven by soil temperature.

For future work, we suggest further exploration of the underlying mechanisms driving the observed relationships between soil temperature and phenology. Investigating the physiological responses of plant species to soil temperature variations and exploring the interactions between soil temperature and other environmental factors at finer temporal and spatial scales would provide a more comprehensive understanding.

In addition, incorporating advanced remote sensing techniques, such as satellite imagery, in conjunction with ground-based measurements can improve the accuracy and comprehensiveness of phenological studies in subarctic grassland ecosystems. Long-term monitoring at multiple sites and the incorporation of various geographical locations would provide valuable information on the generalizability of our findings and the response of subarctic grasslands to ongoing climate change.

This study contributes to our knowledge of the relationships between soil temperature, other meteorological variables, and vegetation phenology in subarctic grassland ecosystems. The findings enhance our understanding of the mechanisms driving ecosystem dynamics in these regions and have implications for predicting and managing subarctic grasslands in the face of environmental change. Finally, this work also functions as a proof-of-concept for \ac{ML}-based vegetation phenology studies, and thereby provides a solid foundation for future research in this domain.

\section{CRediT author statement}

\textbf{Steven Mortier:} Conceptualization, Methodology, Software, Validation, Formal analysis, Data Curation, Writing - Original Draft, Visualization. 
\textbf{Amir Hamedpour:} Conceptualization, Investigation, Writing - Original Draft. \textbf{Bart Bussmann:} Conceptualization, Methodology, Software, Data Curation. 
\textbf{Ruth Phoebe Tchana Wandji:} Conceptualization, Investigation, Writing - Original Draft. 
\textbf{Steven Latré:} Funding acquisition, Supervision. 
\textbf{Bjarni Diðrik Sigurdsson:} Investigation, Writing - Review \& Editing, Funding acquisition, Supervision, Project administration. 
\textbf{Tom De Schepper:} Conceptualization, Methodology, Writing - Review \& Editing, Supervision.
\textbf{Tim Verdonck:} Conceptualization, Methodology, Resources, Writing - Review \& Editing, Supervision.

\section{Acknowledgements}

Funding: This work was supported by the European Union’s Horizon 2020 research and innovation program under the Marie Sklodowska-Curie grant agreement No 813114.

\bibliographystyle{plainnat}
\bibliography{references_semimanual}

\begin{thebibliography}{77}
\providecommand{\natexlab}[1]{#1}
\providecommand{\url}[1]{\texttt{#1}}
\expandafter\ifx\csname urlstyle\endcsname\relax
  \providecommand{\doi}[1]{doi: #1}\else
  \providecommand{\doi}{doi: \begingroup \urlstyle{rm}\Url}\fi

\bibitem[A.~S.~Hope and Stow(1993)]{hope1993theRelationship}
J.~S.~Kimball A.~S.~Hope and D.~A. Stow.
\newblock The relationship between tussock tundra spectral reflectance properties and biomass and vegetation composition.
\newblock \emph{International Journal of Remote Sensing}, 14\penalty0 (10):\penalty0 1861--1874, 1993.
\newblock \doi{10.1080/01431169308954008}.

\bibitem[Aas et~al.(2021)Aas, Jullum, and Løland]{AAS2021103502}
Kjersti Aas, Martin Jullum, and Anders Løland.
\newblock Explaining individual predictions when features are dependent: More accurate approximations to shapley values.
\newblock \emph{Artificial Intelligence}, 298:\penalty0 103502, 2021.
\newblock ISSN 0004-3702.
\newblock \doi{https://doi.org/10.1016/j.artint.2021.103502}.

\bibitem[Akiba et~al.(2019)Akiba, Sano, Yanase, Ohta, and Koyama]{Akiba2019Optuna:Framework}
Takuya Akiba, Shotaro Sano, Toshihiko Yanase, Takeru Ohta, and Masanori Koyama.
\newblock {Optuna: A Next-generation Hyperparameter Optimization Framework}.
\newblock \emph{Proceedings of the ACM SIGKDD International Conference on Knowledge Discovery and Data Mining}, pages 2623--2631, 7 2019.
\newblock \doi{10.1145/3292500.3330701}.

\bibitem[Arndt et~al.(2019)Arndt, Santos, Ustin, Davidson, Stow, Oechel, Tran, Graybill, and Zona]{Arndt2019ArcticAlaska}
Kyle~A. Arndt, Maria~J. Santos, Susan Ustin, Scott~J. Davidson, Doug Stow, Walter~C. Oechel, Thao~T.P. Tran, Brian Graybill, and Donatella Zona.
\newblock {Arctic greening associated with lengthening growing seasons in Northern Alaska}.
\newblock \emph{Environmental Research Letters}, 14\penalty0 (12):\penalty0 125018, 2019.
\newblock ISSN 17489326.
\newblock \doi{10.1088/1748-9326/ab5e26}.

\bibitem[Balzarolo et~al.(2011)Balzarolo, Anderson, Nichol, Rossini, Vescovo, Arriga, Wohlfahrt, Calvet, Carrara, Cerasoli, Cogliati, Daumard, Eklundh, Elbers, Evrendilek, Handcock, Kaduk, Klumpp, Longdoz, Matteucci, Meroni, Montagnani, Ourcival, S{\'{a}}nchez-Ca{\~{n}}ete, Pontailler, Juszczak, Scholes, and Mart{\'{i}}n]{Balzarolo2011Ground-BasedControversies}
Manuela Balzarolo, Karen Anderson, Caroline Nichol, Micol Rossini, Loris Vescovo, Nicola Arriga, Georg Wohlfahrt, Jean-Christophe Calvet, Arnaud Carrara, Sofia Cerasoli, Sergio Cogliati, Fabrice Daumard, Lars Eklundh, Jan~A. Elbers, Fatih Evrendilek, Rebecca~N. Handcock, Jörg Kaduk, Katja Klumpp, Bernard Longdoz, Giorgio Matteucci, Michele Meroni, Lenoardo Montagnani, Jean-Marc Ourcival, Enrique~P. S{\'{a}}nchez-Ca{\~{n}}ete, Jean-Yves Pontailler, Radoslaw Juszczak, Bob Scholes, and M.~Pilar Mart{\'{i}}n.
\newblock {Ground-Based Optical Measurements at European Flux Sites: A Review of Methods, Instruments and Current Controversies}.
\newblock \emph{Sensors}, 11\penalty0 (8):\penalty0 7954--7981, 2011.
\newblock ISSN 1424-8220.
\newblock \doi{10.3390/s110807954}.

\bibitem[Barr{\'{e}} et~al.(2017)Barr{\'{e}}, St{\"{o}}ver, M{\"{u}}ller, and Steinhage]{Barre2017LeafNet:Identification}
Pierre Barr{\'{e}}, Ben~C. St{\"{o}}ver, Kai~F. M{\"{u}}ller, and Volker Steinhage.
\newblock {LeafNet: A computer vision system for automatic plant species identification}.
\newblock \emph{Ecological Informatics}, 40:\penalty0 50--56, 7 2017.
\newblock ISSN 1574-9541.
\newblock \doi{10.1016/J.ECOINF.2017.05.005}.

\bibitem[Beck and Goetz(2011)]{Beck2011SatelliteDifferences}
Pieter~S.A. Beck and Scott~J. Goetz.
\newblock {Satellite observations of high northern latitude vegetation productivity changes between 1982 and 2008: ecological variability and regional differences}.
\newblock \emph{Environmental Research Letters}, 6\penalty0 (4):\penalty0 045501, 10 2011.
\newblock ISSN 1748-9326.
\newblock \doi{10.1088/1748-9326/6/4/045501}.

\bibitem[Beck et~al.(2006)Beck, Atzberger, H{\o}gda, Johansen, and Skidmore]{Beck2006ImprovedNDVI}
Pieter~S.A. Beck, Clement Atzberger, Kjell~Arild H{\o}gda, Bernt Johansen, and Andrew~K. Skidmore.
\newblock {Improved monitoring of vegetation dynamics at very high latitudes: A new method using MODIS NDVI}.
\newblock \emph{Remote Sensing of Environment}, 100\penalty0 (3):\penalty0 321--334, 2006.
\newblock ISSN 00344257.
\newblock \doi{10.1016/j.rse.2005.10.021}.

\bibitem[Beer et~al.(2018)Beer, Porada, Ekici, and Brakebusch]{Beer2018EffectsRegions}
C~Beer, P~Porada, A~Ekici, and M~Brakebusch.
\newblock {Effects of short-term variability of meteorological variables on soil temperature in permafrost regions}.
\newblock \emph{The Cryosphere}, 12\penalty0 (2):\penalty0 741--757, 2018.
\newblock \doi{10.5194/tc-12-741-2018}.

\bibitem[Bhatt et~al.(2013)Bhatt, Walker, Raynolds, Bieniek, Epstein, Comiso, Pinzon, Tucker, and Polyakov]{Bhatt2013RecentTundra}
Uma~S. Bhatt, Donald~A. Walker, Martha~K. Raynolds, Peter~A. Bieniek, Howard~E. Epstein, Josefino~C. Comiso, Jorge~E. Pinzon, Compton~J. Tucker, and Igor~V. Polyakov.
\newblock {Recent Declines in Warming and Vegetation Greening Trends over Pan-Arctic Tundra}.
\newblock \emph{Remote Sensing 2013, Vol. 5, Pages 4229-4254}, 5\penalty0 (9):\penalty0 4229--4254, 8 2013.
\newblock ISSN 2072-4292.
\newblock \doi{10.3390/RS5094229}.

\bibitem[Bjorkman et~al.(2020)Bjorkman, Garc{\'{i}}a~Criado, Myers-Smith, Ravolainen, J{\'{o}}nsd{\'{o}}ttir, Westergaard, Lawler, Aronsson, Bennett, Gardfjell, Heiðmarsson, Stewart, and Normand]{Bjorkman2020StatusMonitoring}
Anne~D. Bjorkman, Mariana Garc{\'{i}}a~Criado, Isla~H. Myers-Smith, Virve Ravolainen, Ingibjörg~Svala J{\'{o}}nsd{\'{o}}ttir, Kristine~Bakke Westergaard, James~P. Lawler, Mora Aronsson, Bruce Bennett, Hans Gardfjell, Starri Heiðmarsson, Laerke Stewart, and Signe Normand.
\newblock {Status and trends in Arctic vegetation: Evidence from experimental warming and long-term monitoring}.
\newblock \emph{Ambio}, 49\penalty0 (3):\penalty0 678--692, 2020.
\newblock ISSN 16547209.
\newblock \doi{10.1007/s13280-019-01161-6}.

\bibitem[Bj{\"{o}}rnsson et~al.(2007)Bj{\"{o}}rnsson, Sigurðsson, Dav{\'{i}}ðsd{\'{o}}ttir, {\'{O}}lafsson, {\'{A}}stþ{\'{o}}rssson, {\'{O}}lafsd{\'{o}}ttir, Baldursson, and J{\'{o}}nsson]{Bjornsson2007Loftslagsbreytingar2018}
Halldór Bj{\"{o}}rnsson, Bjarni~D. Sigurðsson, Brynhildur Dav{\'{i}}ðsd{\'{o}}ttir, Jón {\'{O}}lafsson, {\'{O}}lafur~S. {\'{A}}stþ{\'{o}}rssson, Snjólaug {\'{O}}lafsd{\'{o}}ttir, Trausti Baldursson, and Trausti J{\'{o}}nsson.
\newblock \emph{{Loftslagsbreytingar og {\'{a}}hrif þeirra {\'{a}} {\'{I}}slandi: Sk{\'{y}}rsla v{\'{i}}sindanefndar um loftlagsbreytingar 2018}}.
\newblock 2007.
\newblock ISBN 9789935941404.
\newblock URL \url{https://orkustofnun.is/gogn/Skyrslur/OS-2007/OS-2007-001.pdf}.

\bibitem[Chen et~al.(2021)Chen, Lantz, Hermosilla, and Wulder]{Chen2021BiophysicalArctic}
Angel Chen, Trevor~C. Lantz, Txomin Hermosilla, and Michael~A. Wulder.
\newblock {Biophysical controls of increased tundra productivity in the western Canadian Arctic}.
\newblock \emph{Remote Sensing of Environment}, 258:\penalty0 112358, 2021.
\newblock ISSN 00344257.
\newblock \doi{10.1016/j.rse.2021.112358}.

\bibitem[Chen and Guestrin(2016)]{Chen2016XGBoost:System}
Tianqi Chen and Carlos Guestrin.
\newblock {XGBoost: A Scalable Tree Boosting System}.
\newblock \emph{Proceedings of the ACM SIGKDD International Conference on Knowledge Discovery and Data Mining}, 13-17-August-2016:\penalty0 785--794, 3 2016.
\newblock \doi{10.1145/2939672.2939785}.

\bibitem[Chen et~al.(2020)Chen, Zhao, Chen, Zhou, and Hughes]{Chen2020AutomaticApproaches}
Xing Chen, Jun Zhao, Yan~hua Chen, Wei Zhou, and Alice~C. Hughes.
\newblock {Automatic standardized processing and identification of tropical bat calls using deep learning approaches}.
\newblock \emph{Biological Conservation}, 241:\penalty0 108269, 1 2020.
\newblock ISSN 0006-3207.
\newblock \doi{10.1016/J.BIOCON.2019.108269}.

\bibitem[Cho et~al.(2009)Cho, Choi, Lee, and Park]{Cho2009CharacterizingMap}
Hee~Sun Cho, Kwang~Hee Choi, Sang~Don Lee, and Young~Seuk Park.
\newblock {Characterizing habitat preference of Eurasian river otter (Lutra lutra) in streams using a self-organizing map}.
\newblock \emph{Limnology}, 10\penalty0 (3):\penalty0 203--213, 6 2009.
\newblock ISSN 14398621.
\newblock \doi{10.1007/S10201-009-0275-7/TABLES/2}.

\bibitem[Christin et~al.(2019)Christin, Hervet, and Lecomte]{Christin2019ApplicationsEcology}
Sylvain Christin, Eric Hervet, and Nicolas Lecomte.
\newblock {Applications for deep learning in ecology}.
\newblock \emph{Methods in Ecology and Evolution}, 10\penalty0 (10):\penalty0 1632--1644, 10 2019.
\newblock ISSN 2041-210X.
\newblock \doi{10.1111/2041-210X.13256}.

\bibitem[Clapham et~al.(2020)Clapham, Miller, Nguyen, and Darimont]{Clapham2020AutomatedBears}
Melanie Clapham, Ed~Miller, Mary Nguyen, and Chris~T. Darimont.
\newblock {Automated facial recognition for wildlife that lack unique markings: A deep learning approach for brown bears}.
\newblock \emph{Ecology and Evolution}, 10\penalty0 (23):\penalty0 12883--12892, 12 2020.
\newblock ISSN 2045-7758.
\newblock \doi{10.1002/ECE3.6840}.

\bibitem[Collins et~al.(2021)Collins, Elmendorf, Hollister, Henry, Clark, Bjorkman, Myers-Smith, Prevéy, Ashton, Assmann, Alatalo, Carbognani, Chisholm, Cooper, Forrester, Jónsdóttir, Klanderud, Kopp, Livensperger, Mauritz, May, Molau, Oberbauer, Ogburn, Panchen, Petraglia, Post, Rixen, Rodenhizer, Schuur, Semenchuk, Smith, Steltzer, Ørjan Totland, Walker, Welker, and Suding]{Collins2021}
Courtney~G Collins, Sarah~C Elmendorf, Robert~D Hollister, Greg H~R Henry, Karin Clark, Anne~D Bjorkman, Isla~H Myers-Smith, Janet~S Prevéy, Isabel~W Ashton, Jakob~J Assmann, Juha~M Alatalo, Michele Carbognani, Chelsea Chisholm, Elisabeth~J Cooper, Chiara Forrester, Ingibjörg~Svala Jónsdóttir, Kari Klanderud, Christopher~W Kopp, Carolyn Livensperger, Marguerite Mauritz, Jeremy~L May, Ulf Molau, Steven~F Oberbauer, Emily Ogburn, Zoe~A Panchen, Alessandro Petraglia, Eric Post, Christian Rixen, Heidi Rodenhizer, Edward A~G Schuur, Philipp Semenchuk, Jane~G Smith, Heidi Steltzer, Ørjan Totland, Marilyn~D Walker, Jeffrey~M Welker, and Katharine~N Suding.
\newblock Experimental warming differentially affects vegetative and reproductive phenology of tundra plants.
\newblock \emph{Nature Communications}, 12:\penalty0 3442, 2021.
\newblock ISSN 2041-1723.
\newblock \doi{10.1038/s41467-021-23841-2}.

\bibitem[Conn et~al.(2000)Conn, Gould, and Toint]{Conn2000TrustMethods}
Andrew~R Conn, Nicholas I~M Gould, and Philippe~L Toint.
\newblock \emph{{Trust region methods}}.
\newblock SIAM, 2000.

\bibitem[Epstein et~al.(2012)Epstein, Raynolds, Walker, Bhatt, Tucker, and Pinzon]{Epstein2012DynamicsDecades}
Howard~E. Epstein, Martha~K. Raynolds, Donald~A. Walker, Uma~S. Bhatt, Compton~J. Tucker, and Jorge~E. Pinzon.
\newblock {Dynamics of aboveground phytomass of the circumpolar Arctic tundra during the past three decades}.
\newblock \emph{Environmental Research Letters}, 7\penalty0 (1), 2012.
\newblock ISSN 17489326.
\newblock \doi{10.1088/1748-9326/7/1/015506}.

\bibitem[Epstein et~al.(2013)Epstein, Myers-Smith, and Walker]{Epstein2013RecentVegetation}
Howard~E. Epstein, Isla Myers-Smith, and Donald~A. Walker.
\newblock {Recent dynamics of arctic and sub-arctic vegetation}.
\newblock \emph{Environmental Research Letters}, 8\penalty0 (1), 2013.
\newblock ISSN 17489326.
\newblock \doi{10.1088/1748-9326/8/1/015040}.

\bibitem[Estrella et~al.(2021)Estrella, Stoeth, Krakauer, and Devineni]{estrella2021quantifying}
E~Herrera Estrella, A~Stoeth, NY~Krakauer, and N~Devineni.
\newblock Quantifying vegetation response to environmental changes on the galapagos islands, ecuador using the normalized difference vegetation index (ndvi).
\newblock \emph{Environmental Research Communications}, 3\penalty0 (6):\penalty0 065003, 2021.
\newblock \doi{10.1088/2515-7620/ac0bd1}.

\bibitem[Fenner(1998)]{Fenner1998ThePlants}
Michael Fenner.
\newblock {The phenology of growth and reproduction in plants}.
\newblock \emph{Perspectives in Plant Ecology, Evolution and Systematics}, 1\penalty0 (1):\penalty0 78--91, 1 1998.
\newblock ISSN 1433-8319.
\newblock \doi{10.1078/1433-8319-00053}.

\bibitem[Ferrara et~al.(2010)Ferrara, Fiorentino, Martinelli, Garofalo, and Rana]{Ferrara2010ComparisonProperties}
Rossana~Monica Ferrara, Costanza Fiorentino, Nicola Martinelli, Pasquale Garofalo, and Gianfranco Rana.
\newblock {Comparison of different ground-based NDVI measurement methodologies to evaluate crop biophysical properties}.
\newblock \emph{Italian Journal of Agronomy}, 5\penalty0 (2):\penalty0 145--154, 2010.
\newblock ISSN 11254718.
\newblock \doi{10.4081/ija.2010.145}.

\bibitem[Frye et~al.(2020)Frye, Rowat, and Feige]{frye2020asymmetric}
Christopher Frye, Colin Rowat, and Ilya Feige.
\newblock Asymmetric shapley values: incorporating causal knowledge into model-agnostic explainability.
\newblock \emph{Advances in Neural Information Processing Systems}, 33:\penalty0 1229--1239, 2020.

\bibitem[Guo et~al.(2020)Guo, Jin, Li, Yang, Xu, Ju, Zhang, Xuan, Liu, Su, Xu, and Liu]{Guo2020ApplicationChallenges}
Qinghua Guo, Shichao Jin, Min Li, Qiuli Yang, Kexin Xu, Yuanzhen Ju, Jing Zhang, Jing Xuan, Jin Liu, Yanjun Su, Qiang Xu, and Yu~Liu.
\newblock {Application of deep learning in ecological resource research: Theories, methods, and challenges}.
\newblock \emph{Science China Earth Sciences}, 63\penalty0 (10):\penalty0 1457--1474, 10 2020.
\newblock ISSN 18691897.
\newblock \doi{10.1007/S11430-019-9584-9}.

\bibitem[He et~al.(2022)He, Zhao, Mao, and Griffin-Nolanb]{He2022ExplainableDistribution}
Bohao He, Yanghe Zhao, Wei Mao, and Robert~J. Griffin-Nolanb.
\newblock {Explainable artificial intelligence reveals environmental constraints in seagrass distribution}.
\newblock \emph{Ecological Indicators}, 144:\penalty0 109523, 11 2022.
\newblock ISSN 1470-160X.
\newblock \doi{10.1016/J.ECOLIND.2022.109523}.

\bibitem[Hearst et~al.(1998)Hearst, Dumais, Osuna, Platt, and Scholkopf]{Hearst1998SupportMachines}
M~A Hearst, S~T Dumais, E~Osuna, J~Platt, and B~Scholkopf.
\newblock {Support vector machines}.
\newblock \emph{IEEE Intelligent Systems and their Applications}, 13\penalty0 (4):\penalty0 18--28, 7 1998.
\newblock ISSN 2374-9423.
\newblock \doi{10.1109/5254.708428}.

\bibitem[Hou et~al.(2015)Hou, Gao, Wu, and Dai]{Hou2015InterannualChina}
Wenjuan Hou, Jiangbo Gao, Shaohong Wu, and Erfu Dai.
\newblock {Interannual variations in growing-season NDVI and its correlation with climate variables in the southwestern karst region of China}.
\newblock \emph{Remote Sensing}, 7\penalty0 (9):\penalty0 11105--11124, 2015.
\newblock ISSN 20724292.
\newblock \doi{10.3390/rs70911105}.

\bibitem[Huang et~al.(2021)Huang, Tang, Hupy, Wang, and Shao]{Huang2021ASensing}
Sha Huang, Lina Tang, Joseph~P Hupy, Yang Wang, and Guofan Shao.
\newblock {A commentary review on the use of normalized difference vegetation index (NDVI) in the era of popular remote sensing}.
\newblock \emph{Journal of Forestry Research}, 32\penalty0 (1):\penalty0 1--6, 2021.
\newblock ISSN 1993-0607.
\newblock \doi{10.1007/s11676-020-01155-1}.

\bibitem[{IPCC}(2021)]{IPCC2021TechnicalChange}
{IPCC}.
\newblock {Technical Summary. Contribution of Working Group I to the Sixth Assessment Report of the Intergovernmental Panel on Climate Change}.
\newblock In V~Masson-Delmotte, P~Zhai, A~Pirani, S~L Connors, C~P{\'{e}}an, S~Berger, N~Caud, Y~Chen, L~Goldfarb, M~I Gomis, M~Huang, K~Leitzell, E~Lonnoy, J~B~R Matthews, T~K Maycock, T~Waterfield, O~Yelek{\c{c}}i, R~Yu, and B~Zhou, editors, \emph{Climate Change 2021: The Physical Science Basis. Contribution of Working Group I to the Sixth Assessment Report of the Intergovernmental Panel on Climate Change}, pages 33--144. Cambridge University Press, Cambridge, United Kingdom and New York, NY, USA, 2021.
\newblock \doi{10.1017/9781009157896.002.}

\bibitem[Karlsen et~al.(2014)Karlsen, Elvebakk, H{\o}gda, and Grydeland]{Karlsen2014SpatialData}
Stein~Rune Karlsen, Arve Elvebakk, Kjell~Arild H{\o}gda, and Tom Grydeland.
\newblock {Spatial and Temporal Variability in the Onset of the Growing Season on Svalbard, Arctic Norway — Measured by MODIS-NDVI Satellite Data}.
\newblock \emph{Remote Sensing}, 6\penalty0 (9):\penalty0 8088--8106, 2014.
\newblock ISSN 2072-4292.
\newblock \doi{10.3390/rs6098088}.

\bibitem[Ke et~al.(2017)Ke, Meng, Finley, Wang, Chen, Ma, Ye, and Liu]{Ke2017LightGBM:Tree}
Guolin Ke, Qi~Meng, Thomas Finley, Taifeng Wang, Wei Chen, Weidong Ma, Qiwei Ye, and Tie-Yan Liu.
\newblock {LightGBM: A Highly Efficient Gradient Boosting Decision Tree}.
\newblock \emph{Advances in Neural Information Processing Systems}, 30, 2017.
\newblock URL \url{https://github.com/Microsoft/LightGBM.}

\bibitem[Leblans et~al.(2017)Leblans, Sigurdsson, Vicca, Fu, Penuelas, and Janssens]{Leblans2017PhenologicalWarming}
Niki~I.W. Leblans, Bjarni~D. Sigurdsson, Sara Vicca, Yongshuo Fu, Josep Penuelas, and Ivan~A. Janssens.
\newblock {Phenological responses of Icelandic subarctic grasslands to short-term and long-term natural soil warming}.
\newblock \emph{Global Change Biology}, 23\penalty0 (11):\penalty0 4932--4945, 2017.
\newblock ISSN 13652486.
\newblock \doi{10.1111/gcb.13749}.

\bibitem[Li et~al.(2020)Li, Buitenwerf, Munk, B{\o}cher, and Svenning]{Li2020Deep-learningEcosystem}
Wang Li, Robert Buitenwerf, Michael Munk, Peder~Klith B{\o}cher, and Jens~Christian Svenning.
\newblock {Deep-learning based high-resolution mapping shows woody vegetation densification in greater Maasai Mara ecosystem}.
\newblock \emph{Remote Sensing of Environment}, 247:\penalty0 111953, 9 2020.
\newblock ISSN 0034-4257.
\newblock \doi{10.1016/J.RSE.2020.111953}.

\bibitem[Loranty and Goetz(2012)]{Loranty2012ShrubTundra}
Michael~M. Loranty and Scott~J. Goetz.
\newblock {Shrub expansion and climate feedbacks in Arctic tundra}.
\newblock \emph{Environmental Research Letters}, 7\penalty0 (1), 2012.
\newblock ISSN 17489326.
\newblock \doi{10.1088/1748-9326/7/1/011005}.

\bibitem[Lumbierres et~al.(2017)Lumbierres, Méndez, Bustamante, Soriguer, and Santamaría]{Lumbierres2017}
Maria Lumbierres, Pablo~F Méndez, Javier Bustamante, Ramón Soriguer, and Luis Santamaría.
\newblock Modeling biomass production in seasonal wetlands using modis ndvi land surface phenology.
\newblock \emph{Remote Sensing}, 9, 2017.
\newblock ISSN 2072-4292.
\newblock \doi{10.3390/rs9040392}.

\bibitem[Lundberg et~al.(2017)Lundberg, Allen, and Lee]{Lundberg2017APredictions}
Scott~M Lundberg, Paul~G Allen, and Su-In Lee.
\newblock {A Unified Approach to Interpreting Model Predictions}.
\newblock \emph{Advances in Neural Information Processing Systems}, 30, 2017.
\newblock URL \url{https://github.com/slundberg/shap}.

\bibitem[Masago and Lian(2022)]{Masago2022EstimatingAlgorithms}
Yoshifumi Masago and Maychee Lian.
\newblock {Estimating the first flowering and full blossom dates of Yoshino cherry (Cerasus × yedoensis ‘Somei-yoshino’) in Japan using machine learning algorithms}.
\newblock \emph{Ecological Informatics}, 71:\penalty0 101835, 11 2022.
\newblock ISSN 1574-9541.
\newblock \doi{10.1016/J.ECOINF.2022.101835}.

\bibitem[McCulloch and Pitts(1943)]{McCulloch1943AActivity}
Warren~S. McCulloch and Walter Pitts.
\newblock {A logical calculus of the ideas immanent in nervous activity}.
\newblock \emph{The Bulletin of Mathematical Biophysics}, 5\penalty0 (4):\penalty0 115--133, 12 1943.
\newblock ISSN 00074985.
\newblock \doi{10.1007/BF02478259/METRICS}.

\bibitem[Merrington(2019)]{Merrington2019AData}
Alexander~Thomas Merrington.
\newblock {A Time Series Analysis of Vegetation Succession on Lava Flow Fields at Hekla Volcano: Assessing the Utility of Landsat Data}, 2019.
\newblock URL \url{https://skemman.is/handle/1946/33203}.

\bibitem[Michielsen(2014)]{Michielsen2014}
Lieven Michielsen.
\newblock Plant communities and global change: adaptation by changes in present species composition or adaptation in plant traits. a case study in iceland.
\newblock Master's thesis, Universiteit Antwerpen, 2014.
\newblock URL \url{https://anet.be/record/opacuantwerpen/c:lvd:14296534/N}.

\bibitem[Molnar et~al.(2020)Molnar, K{\"o}nig, Herbinger, Freiesleben, Dandl, Scholbeck, Casalicchio, Grosse-Wentrup, and Bischl]{Molnar2021GeneralModels}
Christoph Molnar, Gunnar K{\"o}nig, Julia Herbinger, Timo Freiesleben, Susanne Dandl, Christian~A Scholbeck, Giuseppe Casalicchio, Moritz Grosse-Wentrup, and Bernd Bischl.
\newblock General pitfalls of model-agnostic interpretation methods for machine learning models.
\newblock In \emph{International Workshop on Extending Explainable AI Beyond Deep Models and Classifiers}, pages 39--68. Springer, 2020.

\bibitem[Myers-Smith et~al.(2020)Myers-Smith, Kerby, Phoenix, Bjerke, Epstein, Assmann, John, Andreu-Hayles, Angers-Blondin, Beck, Berner, Bhatt, Bjorkman, Blok, Bryn, Christiansen, Cornelissen, Cunliffe, Elmendorf, Forbes, Goetz, Hollister, de~Jong, Loranty, Macias-Fauria, Maseyk, Normand, Olofsson, Parker, Parmentier, Post, Schaepman-Strub, Stordal, Sullivan, Thomas, T{\o}mmervik, Treharne, Tweedie, Walker, Wilmking, and Wipf]{Myers-Smith2020ComplexityArctic}
Isla~H. Myers-Smith, Jeffrey~T. Kerby, Gareth~K. Phoenix, Jarle~W. Bjerke, Howard~E. Epstein, Jakob~J. Assmann, Christian John, Laia Andreu-Hayles, Sandra Angers-Blondin, Pieter~S.A. Beck, Logan~T. Berner, Uma~S. Bhatt, Anne~D. Bjorkman, Daan Blok, Anders Bryn, Casper~T. Christiansen, J.~Hans~C. Cornelissen, Andrew~M. Cunliffe, Sarah~C. Elmendorf, Bruce~C. Forbes, Scott~J. Goetz, Robert~D. Hollister, Rogier de~Jong, Michael~M. Loranty, Marc Macias-Fauria, Kadmiel Maseyk, Signe Normand, Johan Olofsson, Thomas~C. Parker, Frans Jan~W. Parmentier, Eric Post, Gabriela Schaepman-Strub, Frode Stordal, Patrick~F. Sullivan, Haydn~J.D. Thomas, Hans T{\o}mmervik, Rachael Treharne, Craig~E. Tweedie, Donald~A. Walker, Martin Wilmking, and Sonja Wipf.
\newblock {Complexity revealed in the greening of the Arctic}.
\newblock \emph{Nature Climate Change 2020 10:2}, 10\penalty0 (2):\penalty0 106--117, 1 2020.
\newblock ISSN 1758-6798.
\newblock \doi{10.1038/s41558-019-0688-1}.

\bibitem[Olafsson and Rousta(2021)]{Olafsson2021InfluenceSensing}
Haraldur Olafsson and Iman Rousta.
\newblock {Influence of atmospheric patterns and North Atlantic Oscillation (NAO) on vegetation dynamics in Iceland using Remote Sensing}.
\newblock \emph{European Journal of Remote Sensing}, 54\penalty0 (1):\penalty0 351--363, 2021.
\newblock ISSN 22797254.
\newblock \doi{10.1080/22797254.2021.1931462}.

\bibitem[O’Gorman and Dwyer(2018)]{OGorman2018UsingEvents}
Paul~A. O’Gorman and John~G. Dwyer.
\newblock {Using Machine Learning to Parameterize Moist Convection: Potential for Modeling of Climate, Climate Change, and Extreme Events}.
\newblock \emph{Journal of Advances in Modeling Earth Systems}, 10\penalty0 (10):\penalty0 2548--2563, 10 2018.
\newblock ISSN 1942-2466.
\newblock \doi{10.1029/2018MS001351}.

\bibitem[Park et~al.(2022)Park, Lee, Kim, Park, Lee, and Heo]{Park2022InterpretationIntelligence}
Jungsu Park, Woo~Hyoung Lee, Keug~Tae Kim, Cheol~Young Park, Sanghun Lee, and Tae~Young Heo.
\newblock {Interpretation of ensemble learning to predict water quality using explainable artificial intelligence}.
\newblock \emph{Science of The Total Environment}, 832:\penalty0 155070, 8 2022.
\newblock ISSN 0048-9697.
\newblock \doi{10.1016/J.SCITOTENV.2022.155070}.

\bibitem[Pedregosa et~al.(2011)Pedregosa, Varoquaux, Gramfort, V., Thirion, Grisel, Blondel, P., Weiss, Dubourg, Vanderplas, Passos, Cournapeau, Brucher, Perrot, and Duchesnay]{Pedregosa2011Scikit-learn:Python}
F~Pedregosa, G~Varoquaux, A~Gramfort, Michel V., B~Thirion, O~Grisel, M~Blondel, Prettenhofer P., R~Weiss, V~Dubourg, J~Vanderplas, A~Passos, D~Cournapeau, M~Brucher, M~Perrot, and E~Duchesnay.
\newblock {Scikit-learn: Machine Learning in Python}.
\newblock \emph{Journal of Machine Learning Research}, 12:\penalty0 2825--2830, 2011.

\bibitem[Perry et~al.(2022)Perry, Sheffield, Crawford, Akpa, Clancy, and Clark]{Perry2022}
Eileen Perry, Kathryn Sheffield, Doug Crawford, Stephen Akpa, Alex Clancy, and Robert Clark.
\newblock Spatial and temporal biomass and growth for grain crops using ndvi time series.
\newblock \emph{Remote Sensing}, 14, 2022.
\newblock ISSN 2072-4292.
\newblock \doi{10.3390/rs14133071}.

\bibitem[Potter and Alexander(2020)]{Potter2020ChangesDecades}
Christopher Potter and Olivia Alexander.
\newblock {Changes in Vegetation Phenology and Productivity in Alaska Over the Past Two Decades}.
\newblock \emph{Remote Sensing}, 12\penalty0 (10), 2020.
\newblock ISSN 2072-4292.
\newblock \doi{10.3390/rs12101546}.

\bibitem[Raynolds et~al.(2015)Raynolds, Magn{\'{u}}sson, Met{\'{u}}salemsson, and Magn{\'{u}}sson]{Raynolds2015WarmingTrends}
Martha Raynolds, Borgthór Magn{\'{u}}sson, Sigmar Met{\'{u}}salemsson, and Sigurdur~H. Magn{\'{u}}sson.
\newblock {Warming, sheep and volcanoes: Land cover changes in Iceland evident in satellite NDVI trends}.
\newblock \emph{Remote Sensing}, 7\penalty0 (8):\penalty0 9492--9506, 2015.
\newblock ISSN 20724292.
\newblock \doi{10.3390/rs70809492}.

\bibitem[Ribeiro et~al.(2016)Ribeiro, Singh, and Guestrin]{Ribeiro2016WhyClassifier}
Marco~Tulio Ribeiro, Sameer Singh, and Carlos Guestrin.
\newblock {"Why should i trust you?" Explaining the predictions of any classifier}.
\newblock \emph{Proceedings of the ACM SIGKDD International Conference on Knowledge Discovery and Data Mining}, 13-17-August-2016:\penalty0 1135--1144, 8 2016.
\newblock \doi{10.1145/2939672.2939778}.

\bibitem[Rolnick et~al.(2022)Rolnick, Donti, Kaack, Kochanski, Lacoste, Sankaran, Ross, Milojevic-Dupont, Jaques, Waldman-Brown, Luccioni, Maharaj, Sherwin, Mukkavilli, Kording, Gomes, Ng, Hassabis, Platt, Creutzig, Chayes, and Bengio]{Rolnick2022TacklingLearning}
David Rolnick, Priya~L. Donti, Lynn~H. Kaack, Kelly Kochanski, Alexandre Lacoste, Kris Sankaran, Andrew~Slavin Ross, Nikola Milojevic-Dupont, Natasha Jaques, Anna Waldman-Brown, Alexandra~Sasha Luccioni, Tegan Maharaj, Evan~D. Sherwin, S.~Karthik Mukkavilli, Konrad~P. Kording, Carla~P. Gomes, Andrew~Y. Ng, Demis Hassabis, John~C. Platt, Felix Creutzig, Jennifer Chayes, and Yoshua Bengio.
\newblock {Tackling Climate Change with Machine Learning}.
\newblock \emph{ACM Computing Surveys (CSUR)}, 55\penalty0 (2):\penalty0 96, 2 2022.
\newblock ISSN 15577341.
\newblock \doi{10.1145/3485128}.

\bibitem[Ryu et~al.(2021)Ryu, Oh, and Cho]{RYU2021SimpleSensor}
Jae~Hyun Ryu, Dohyeok Oh, and Jaeil Cho.
\newblock {Simple method for extracting the seasonal signals of photochemical reflectance index and normalized difference vegetation index measured using a spectral reflectance sensor}.
\newblock \emph{Journal of Integrative Agriculture}, 20\penalty0 (7):\penalty0 1969--1986, 2021.
\newblock ISSN 20953119.
\newblock \doi{10.1016/S2095-3119(20)63410-4}.

\bibitem[Schofield et~al.(2019)Schofield, Nagrani, Zisserman, Hayashi, Matsuzawa, Biro, and Carvalho]{Schofield2019ChimpanzeeLearning}
Daniel Schofield, Arsha Nagrani, Andrew Zisserman, Misato Hayashi, Tetsuro Matsuzawa, Dora Biro, and Susana Carvalho.
\newblock {Chimpanzee face recognition from videos in the wild using deep learning}.
\newblock \emph{Science Advances}, 5\penalty0 (9), 9 2019.
\newblock ISSN 23752548.
\newblock \doi{10.1126/SCIADV.AAW0736}.

\bibitem[Seabold and Perktold(2010)]{Seabold2010Statsmodels:Python}
Skipper Seabold and Josef Perktold.
\newblock {statsmodels: Econometric and statistical modeling with python}.
\newblock In \emph{9th Python in Science Conference}, 2010.

\bibitem[Semenchuk et~al.(2016)Semenchuk, Gillespie, Rumpf, Baggesen, Elberling, and Cooper]{Semenchuk2016HighPeriodicity}
Philipp~R. Semenchuk, Mark~A.K. Gillespie, Sabine~B. Rumpf, Nanna Baggesen, Bo~Elberling, and Elisabeth~J. Cooper.
\newblock {High Arctic plant phenology is determined by snowmelt patterns but duration of phenological periods is fixed: An example of periodicity}.
\newblock \emph{Environmental Research Letters}, 11\penalty0 (12), 2016.
\newblock ISSN 17489326.
\newblock \doi{10.1088/1748-9326/11/12/125006}.

\bibitem[Sigurdsson(2001)]{Sigurdsson2001ElevatedStudy}
Bjarni~D Sigurdsson.
\newblock {Elevated [CO2] and nutrient status modified leaf phenology and growth rhythm of young Populus trichocarpa trees in a 3-year field study}.
\newblock \emph{Trees}, 15\penalty0 (7):\penalty0 403--413, 2001.
\newblock ISSN 1432-2285.
\newblock \doi{10.1007/s004680100121}.

\bibitem[Sigurdsson et~al.(2016)Sigurdsson, Leblans, Dauwe, Gudmundsd{\'{o}}ttir, Gundersen, Gunnarsd{\'{o}}ttir, Holmstrup, Ilieva-Makulec, K{\"{a}}tterer, Marteinsd{\'{o}}ttir, Maljanen, Oddsd{\'{o}}ttir, Ostonen, Pe{\~{n}}uelas, Poeplau, Richter, Sigurdsson, Van~Bodegom, Wallander, Weedon, and Janssens]{Sigurdsson2016GeothermalStudy}
Bjarni~D. Sigurdsson, Niki~I.W. Leblans, Steven Dauwe, Elín Gudmundsd{\'{o}}ttir, Per Gundersen, Gunnhildur~E. Gunnarsd{\'{o}}ttir, Martin Holmstrup, Krassimira Ilieva-Makulec, Thomas K{\"{a}}tterer, Bryndís Marteinsd{\'{o}}ttir, Marja Maljanen, Edda~S. Oddsd{\'{o}}ttir, Ivika Ostonen, Josep Pe{\~{n}}uelas, Christopher Poeplau, Andreas Richter, Páll Sigurdsson, Peter Van~Bodegom, Håkan Wallander, James Weedon, and Ivan Janssens.
\newblock {Geothermal ecosystems as natural climate change experiments: The ForHot research site in Iceland as a case study}.
\newblock \emph{Icelandic Agricultural Sciences}, 29\penalty0 (1):\penalty0 53--71, 2016.
\newblock ISSN 1670567X.
\newblock \doi{10.16886/IAS.2016.05}.

\bibitem[Street and Caldararu(2022)]{Street2022WhyLimited}
Lorna~E. Street and S.~Caldararu.
\newblock {Why are Arctic shrubs becoming more nitrogen limited?}
\newblock \emph{New Phytologist}, 233\penalty0 (2):\penalty0 585--587, 1 2022.
\newblock ISSN 1469-8137.
\newblock \doi{10.1111/NPH.17841}.

\bibitem[Strydom et~al.(2021)Strydom, Catchen, Banville, Caron, Dansereau, Desjardins-Proulx, Forero-Mu{\~{n}}oz, Higino, Mercier, Gonzalez, Gravel, Pollock, and Poisot]{Strydom2021ATime}
Tanya Strydom, Michael~D. Catchen, Francis Banville, Dominique Caron, Gabriel Dansereau, Philippe Desjardins-Proulx, Norma~R. Forero-Mu{\~{n}}oz, Gracielle Higino, Benjamin Mercier, Andrew Gonzalez, Dominique Gravel, Laura Pollock, and Timothée Poisot.
\newblock {A roadmap towards predicting species interaction networks (across space and time)}.
\newblock \emph{Philosophical Transactions of the Royal Society B: Biological Sciences}, 376\penalty0 (1837):\penalty0 20210063, 11 2021.
\newblock ISSN 0962-8436.
\newblock \doi{10.1098/rstb.2021.0063}.

\bibitem[Tan et~al.(2022)Tan, Luo, Li, Hao, Wang, Dong, and Chen]{Tan2022InvestigatingChina}
Xiaoqing Tan, Siqiong Luo, Hongmei Li, Xiaohua Hao, Jingyuan Wang, Qingxue Dong, and Zihang Chen.
\newblock {Investigating the Effects of Snow Cover and Vegetation on Soil Temperature Using Remote Sensing Indicators in the Three River Source Region, China}.
\newblock \emph{Remote Sensing}, 14\penalty0 (16), 2022.
\newblock ISSN 2072-4292.
\newblock \doi{10.3390/rs14164114}.

\bibitem[Thessen(2016)]{Thessen2016AdoptionScience}
Anne~E. Thessen.
\newblock {Adoption of Machine Learning Techniques in Ecology and Earth Science}.
\newblock \emph{One Ecosystem 1: e8621}, 1:\penalty0 e8621--, 2016.
\newblock ISSN 2367-8194.
\newblock \doi{10.3897/ONEECO.1.E8621}.

\bibitem[Van Der~Wal and Stien(2014)]{VanDerWal2014High-arcticBiomass}
René Van Der~Wal and Audun Stien.
\newblock {High-arctic plants like it hot: a long-term investigation of between-year variability in plant biomass}.
\newblock \emph{Ecology}, 95\penalty0 (12):\penalty0 3414--3427, 12 2014.
\newblock ISSN 1939-9170.
\newblock \doi{10.1890/14-0533.1}.

\bibitem[Verbrigghe et~al.(2022)Verbrigghe, Leblans, Sigurdsson, Vicca, Fang, Fuchslueger, Soong, Weedon, Poeplau, Ariza-Carricondo, Bahn, Guenet, Gundersen, Gunnarsd{\'{o}}ttir, K{\"{a}}tterer, Liu, Maljanen, Mara{\~{n}}{\'{o}}n-Jim{\'{e}}nez, Meeran, Oddsd{\'{o}}ttir, Ostonen, Pe{\~{n}}uelas, Richter, Sardans, Sigurðsson, Torn, Van~Bodegom, Verbruggen, Walker, Wallander, and Janssens]{Verbrigghe2022SoilTopsoil}
Niel Verbrigghe, Niki~I.W. Leblans, Bjarni~D. Sigurdsson, Sara Vicca, Chao Fang, Lucia Fuchslueger, Jennifer~L. Soong, James~T. Weedon, Christopher Poeplau, Cristina Ariza-Carricondo, Michael Bahn, Bertrand Guenet, Per Gundersen, Gunnhildur~E. Gunnarsd{\'{o}}ttir, Thomas K{\"{a}}tterer, Zhanfeng Liu, Marja Maljanen, Sara Mara{\~{n}}{\'{o}}n-Jim{\'{e}}nez, Kathiravan Meeran, Edda~S. Oddsd{\'{o}}ttir, Ivika Ostonen, Josep Pe{\~{n}}uelas, Andreas Richter, Jordi Sardans, Páll Sigurðsson, Margaret~S. Torn, Peter~M. Van~Bodegom, Erik Verbruggen, Tom~W.N. Walker, Håkan Wallander, and Ivan~A. Janssens.
\newblock {Soil carbon loss in warmed subarctic grasslands is rapid and restricted to topsoil}.
\newblock \emph{Biogeosciences}, 19\penalty0 (14):\penalty0 3381--3393, 2022.
\newblock ISSN 17264189.
\newblock \doi{10.5194/bg-19-3381-2022}.

\bibitem[Vilone and Longo(2021)]{Vilone2021NotionsIntelligence}
Giulia Vilone and Luca Longo.
\newblock {Notions of explainability and evaluation approaches for explainable artificial intelligence}.
\newblock \emph{Information Fusion}, 76:\penalty0 89--106, 12 2021.
\newblock ISSN 1566-2535.
\newblock \doi{10.1016/J.INFFUS.2021.05.009}.

\bibitem[W{\"{a}}ldchen and M{\"{a}}der(2018)]{Waldchen2018MachineIdentification}
Jana W{\"{a}}ldchen and Patrick M{\"{a}}der.
\newblock {Machine learning for image based species identification}.
\newblock \emph{Methods in Ecology and Evolution}, 9\penalty0 (11):\penalty0 2216--2225, 11 2018.
\newblock ISSN 2041-210X.
\newblock \doi{10.1111/2041-210X.13075}.

\bibitem[Walker et~al.(2012{\natexlab{a}})Walker, Epstein, Raynolds, Kuss, Kopecky, Frost, Danils, Leibman, Moskalenko, Matyshak, Khitun, Khomutov, Forbes, Bhatt, Kade, Vonlanthen, and Tich{\'{y}}]{Walker2012EnvironmentTransects}
D.~A. Walker, H.~E. Epstein, M.~K. Raynolds, P.~Kuss, M.~A. Kopecky, G.~V. Frost, F.~J.A. Danils, M.~O. Leibman, N.~G. Moskalenko, G.~V. Matyshak, O.~V. Khitun, A.~V. Khomutov, B.~C. Forbes, U.~S. Bhatt, A.~N. Kade, C.~M. Vonlanthen, and L.~Tich{\'{y}}.
\newblock {Environment, vegetation and greenness (NDVI) along the North America and Eurasia Arctic transects}.
\newblock \emph{Environmental Research Letters}, 7\penalty0 (1), 2012{\natexlab{a}}.
\newblock ISSN 17489326.
\newblock \doi{10.1088/1748-9326/7/1/015504}.

\bibitem[Walker et~al.(2012{\natexlab{b}})Walker, Epstein, Raynolds, Kuss, Kopecky, Frost, Dani{\"e}ls, Leibman, Moskalenko, Matyshak, et~al.]{walker2012environment}
DA~Walker, HE~Epstein, MK~Raynolds, P~Kuss, MA~Kopecky, GV~Frost, FJA Dani{\"e}ls, MO~Leibman, NG~Moskalenko, GV~Matyshak, et~al.
\newblock Environment, vegetation and greenness (ndvi) along the north america and eurasia arctic transects.
\newblock \emph{Environmental Research Letters}, 7\penalty0 (1):\penalty0 015504, 2012{\natexlab{b}}.

\bibitem[Wang et~al.(2021)Wang, Liu, Huang, Bi, Ma, Ma, Shangguan, Zhao, Feng, Liang, Cao, Schmid, and He]{wang2021Satellite}
Hao Wang, Huiying Liu, Ni~Huang, Jian Bi, Xuanlong Ma, Zhiyuan Ma, Zijian Shangguan, Hongfang Zhao, Qisheng Feng, Tiangang Liang, Guangmin Cao, Bernhard Schmid, and Jin-Sheng He.
\newblock Satellite-derived ndvi underestimates the advancement of alpine vegetation growth over the past three decades.
\newblock \emph{Ecology}, 102\penalty0 (12):\penalty0 e03518, 2021.
\newblock \doi{https://doi.org/10.1002/ecy.3518}.

\bibitem[Xie et~al.(2021)Xie, H{\"{u}}sler, de~Jong, Chimani, Asam, Sun, Schaepman, and Kneub{\"{u}}hler]{Xie2021SpringAlps}
Jing Xie, Fabia H{\"{u}}sler, Rogier de~Jong, Barbara Chimani, Sarah Asam, Yeran Sun, Michael~E Schaepman, and Mathias Kneub{\"{u}}hler.
\newblock {Spring Temperature and Snow Cover Climatology Drive the Advanced Springtime Phenology (1991–2014) in the European Alps}.
\newblock \emph{Journal of Geophysical Research: Biogeosciences}, 126\penalty0 (3):\penalty0 e2020JG006150, 2021.
\newblock \doi{https://doi.org/10.1029/2020JG006150}.

\bibitem[Ye and Cai(2011)]{Ye2011ForecastingNetwork}
Lin Ye and Qinghua Cai.
\newblock {Forecasting Daily Chlorophyll a Concentration during the Spring Phytoplankton Bloom Period in Xiangxi Bay of the Three-Gorges Reservoir by Means of a Recurrent Artificial Neural Network}.
\newblock \emph{Journal of Freshwater Ecology}, 24\penalty0 (4):\penalty0 609--617, 2011.
\newblock ISSN 21566941.
\newblock \doi{10.1080/02705060.2009.9664338}.

\bibitem[Zeiler and Fergus(2014)]{Zeiler2014VisualizingNetworks}
Matthew~D. Zeiler and Rob Fergus.
\newblock {Visualizing and Understanding Convolutional Networks}.
\newblock \emph{Computer Vision–ECCV 2014}, 8689\penalty0 (PART 1):\penalty0 818--833, 2014.
\newblock ISSN 978-3-319-10589-5.
\newblock \doi{10.1007/978-3-319-10590-1{\_}53}.

\bibitem[Zhang et~al.(2016)Zhang, Zhang, Xie, Yin, Liu, and Liu]{Zhang2016}
Binghua Zhang, Li~Zhang, Dong Xie, Xiaoli Yin, Chunjing Liu, and Guang Liu.
\newblock Application of synthetic ndvi time series blended from landsat and modis data for grassland biomass estimation.
\newblock \emph{Remote Sensing}, 8, 2016.
\newblock ISSN 2072-4292.
\newblock \doi{10.3390/rs8010010}.

\bibitem[Zhang et~al.(2003)Zhang, Friedl, Schaaf, Strahler, Hodges, Gao, Reed, and Huete]{Zhang2003MonitoringMODIS}
Xiaoyang Zhang, Mark~A. Friedl, Crystal~B. Schaaf, Alan~H. Strahler, John~C.F. Hodges, Feng Gao, Bradley~C. Reed, and Alfredo Huete.
\newblock {Monitoring vegetation phenology using MODIS}.
\newblock \emph{Remote Sensing of Environment}, 84\penalty0 (3):\penalty0 471--475, 3 2003.
\newblock ISSN 0034-4257.
\newblock \doi{10.1016/S0034-4257(02)00135-9}.

\bibitem[Zmarz et~al.(2018)Zmarz, Rodzewicz, D{\c{a}}bski, Karsznia, Korczak-Abshire, and Chwedorzewska]{Zmarz2018ApplicationEcosystem}
Anna Zmarz, Mirosław Rodzewicz, Maciej D{\c{a}}bski, Izabela Karsznia, Małgorzata Korczak-Abshire, and Katarzyna~J Chwedorzewska.
\newblock {Application of UAV BVLOS remote sensing data for multi-faceted analysis of Antarctic ecosystem}.
\newblock \emph{Remote Sensing of Environment}, 217:\penalty0 375--388, 2018.
\newblock ISSN 0034-4257.
\newblock \doi{https://doi.org/10.1016/j.rse.2018.08.031}.

\end{thebibliography}

\end{document}